\documentclass[10pt,journal,compsoc]{IEEEtran}
\usepackage[nocompress]{cite}
\usepackage[pdftex]{graphicx}
\usepackage{amsmath}
\usepackage{algorithmic}
\usepackage{fixltx2e}
\usepackage{stfloats}
\usepackage{comment}
\ifCLASSOPTIONcaptionsoff
  \usepackage[nomarkers]{endfloat}
 \let\MYoriglatexcaption\caption
 \renewcommand{\caption}[2][\relax]{\MYoriglatexcaption[#2]{#2}}
\fi
\usepackage{url}

\usepackage{amssymb}
\usepackage{amsthm}
\usepackage{algorithm}
\usepackage{array}
\usepackage{relsize}
\usepackage{wrapfig}
\usepackage{mathtools}
\usepackage{bm}
\usepackage{calc} 
\usepackage{enumitem}
\usepackage[labelformat=simple]{subfig}
\usepackage{multirow}
\usepackage{booktabs}
\usepackage{xcolor}
\usepackage{hyperref}

\newcommand{\eg}{e.g.}
\newcommand{\ie}{i.e.}
\newcommand{\etal}{et al.}

\newcolumntype{L}[1]{>{\raggedright\let\newline\\\arraybackslash\hspace{0pt}}m{#1}}
\newcolumntype{C}[1]{>{\centering\let\newline\\\arraybackslash\hspace{0pt}}m{#1}}

\begin{document}
\title{An Intermediate-level Attack Framework on The Basis of Linear Regression}

\author{Yiwen Guo, Qizhang Li, Wangmeng Zuo, Hao Chen
  \IEEEcompsocitemizethanks{
  \IEEEcompsocthanksitem Y. Guo is with ByteDance AI Lab. E-mail: guoyiwen89@gmail.com.
  \IEEEcompsocthanksitem Q. Li is with Tencent Security Big Data Lab, and the School of Computer Science and Technology, Harbin Institute of Technology, China. E-mail: liqizhang95@gmail.com.
  \IEEEcompsocthanksitem W. Zuo is with the School of Computer Science and Technology, Harbin Institute of Technology, China. E-mail: cswmzuo@gmail.com.
  \IEEEcompsocthanksitem H. Chen is with the Department of Computer Science, University of California, Davis, US. Email: chen@ucdavis.edu.}
 \thanks{The first two authors contribute equally to this work.}
 }



\IEEEtitleabstractindextext{
 \begin{abstract}
This paper substantially extends our work published at ECCV~\cite{li2020yet}, in which an intermediate-level attack was proposed to improve the transferability of some baseline adversarial examples. Specifically, we advocate a framework in which a direct linear mapping from the intermediate-level discrepancies (between adversarial features and benign features) to prediction loss of the adversarial example is established. By delving deep into the core components of such a framework, we show that 1) a variety of linear regression models can all be considered in order to establish the mapping, 2) the magnitude of the finally obtained intermediate-level adversarial discrepancy is correlated with the transferability, 3) further boost of the performance can be achieved by performing multiple runs of the baseline attack with random initialization. In addition, by leveraging these findings, we achieve new state-of-the-arts on transfer-based $\ell_\infty$ and $\ell_2$ attacks.
Our code is publicly available at \href{https://github.com/qizhangli/ila-plus-plus-lr}{https://github.com/qizhangli/ila-plus-plus-lr}.
 \end{abstract}

 }

\maketitle
\IEEEdisplaynontitleabstractindextext
\IEEEpeerreviewmaketitle

\IEEEraisesectionheading{\section{Introduction}\label{sec:intro}}
\IEEEPARstart{T}{he} community has witnessed a great surge of studies on adversarial examples. It has been shown that, even in a black-box setting where no information about the victim machine learning model was available, an attacker is still capable of generating adversarial examples with reasonably high attack success rates. In general, adversarial attacks in black-box settings are performed via zeroth-order optimization~\cite{Chen2017, Tu2019, Bhagoji2018, Ilyas2018, Ilyas2019, Guo2019}, or applying boundary search~\cite{brendel2017decision, croce2020minimally, chen2020hopskipjumpattack, yan2020policy}, or based on the transferability of adversarial examples~\cite{Papernot2017, Liu2017, Huang2019black, Zhou2018, Yan2019, li2020yet, guo2020back}.

The transferability of adversarial examples has drawn great interest from researchers, owing to its essential role in generating a variety of black-box adversarial examples. The phenomenon shows that adversarial examples generated on one classification model (\ie, a source model) can successfully fool other classification models (\ie, the victim models) which probably have different architectures and different parameters.
Much recent work has been published to shed light on the transferability of adversarial examples and to develop ways of improving it~\cite{Liu2017, Huang2019black, Zhou2018, Yan2019, li2020yet, guo2020back}. In this paper, we also take a step towards explaining and exploiting the adversarial transferability. 

We focus on adversarial examples crafted on the basis of images, which are used for attacking image classification systems or other related computer vision systems. We consider the scenario in which the attacker being given a normal image classification source model $f:\mathbb R^n\times\mathbb C\rightarrow\mathbb R$, which is a deep neural network (DNN) that maps any input image $\mathbf x\in\mathbb R^n$ along with its ground-truth label $y\in\mathbb C$ to evaluation loss, as a composition of a feature extractor $g$ and a classifier $h$ that evaluates the prediction made based on the extracted features, \ie, $f=h\circ g$~\cite{Huang2019black, guo2020back, li2020yet}.
In general, $g$ and $h$ can be considered as two sub-nets of the original DNN model, consisting of its lower and higher layers, respectively. Note that $g$ does not necessarily contains more layers than that of $h$, and we make the depth of $g$ a hyper-parameter that can be tuned.

In this paper, we concern a problem setting (illustrated in Figure~\ref{fig:overview}) in which some initial adversarial examples have already been crafted following a baseline method (\eg, I-FGSM) on a white-box source model, and we aim to enhance their transferability to unknown victim models.
To achieve this, we advocate a framework where a linear mapping (denoted by $h'$) is adopted instead of the original nonlinear one, \ie, $h$, for evaluating the middle-layer features extracted by $g$ which is believed to be largely shared among different DNN models. 
Results of the baseline method are utilized to establish the linear mapping $h'$, and the more transferable adversarial example is anticipated by maximizing the output of $f'=h'\circ g$.

\textbf{What's new in comparison to~\cite{li2020yet}:} 1) By introducing the framework based on linear regression, we show that various models can be applied for obtaining the linear function $h'$, including ridge regression, support vector regression~\cite{platt1999probabilistic}, and ElasticNet~\cite{zou2005regularization}. 2) We provide ample evidence of why (and how) improved adversarial transferability are achieved in such a framework. 3) We show how to further boost the performance of adversarial transferability, in addition to the results in~\cite{li2020yet}. 4) New state-of-the-arts are achieved.


\section{Related Work}

The transferability of adversarial examples has been intensively studied, since it was first discovered~\cite{Szegedy2014}. Ensemble learning~\cite{Liu2017, dong2019evading} and random augmentation~\cite{Xie2019} are two popular strategies for improving the transferability. A series of recent methods also exploit the architecture of advanced DNNs. For instance, Zhou \etal~\cite{Zhou2018} proposed TAP, in which the discrepancy between benign DNN features and their adversarial counterparts was maximized. Similarly, Inkawhich \etal~\cite{Inkawhich2019} encouraged the intermediate-level representation of an adversarial example to be similar to that of a target example. 
Wu \etal~\cite{Wu2020} showed that, for models with skip connections, propagating less gradient through the main stream of a DNN led to more transferable adversarial examples. Guo \etal~\cite{guo2020back} advocated to perform backpropagation without caring about ReLU during backward pass. Other effective methods include~\cite{wang2020unified, wang2021feature, zhu2021rethinking, springer2021little}.

The most relevant work to ours is introduced by Huang \etal~\cite{Huang2019}, since we both follow the problem setting in Figure~\ref{fig:overview}, and we will introduce and discuss it carefully in Section~\ref{sec:ila}.

\section{Linearization and Transferability}\label{sec:linear}

It was originally proposed in our paper~\cite{li2020yet} that the transferability of an off-the-shelf multi-step attack (\eg, the iterative fast gradient sign method, I-FGSM~\cite{Kurakin2017}) can be enhanced by encouraging middle-layer features of the source model to be adversarial, in the sense of maximizing ridge regression loss. The work takes an attempt towards connecting the transferability and linear regression. In what follows, we will first introduce the method and then give more discussions and insights in an unified framework.

\begin{figure}[ht]
    \centering
    \includegraphics[width=0.49\textwidth]{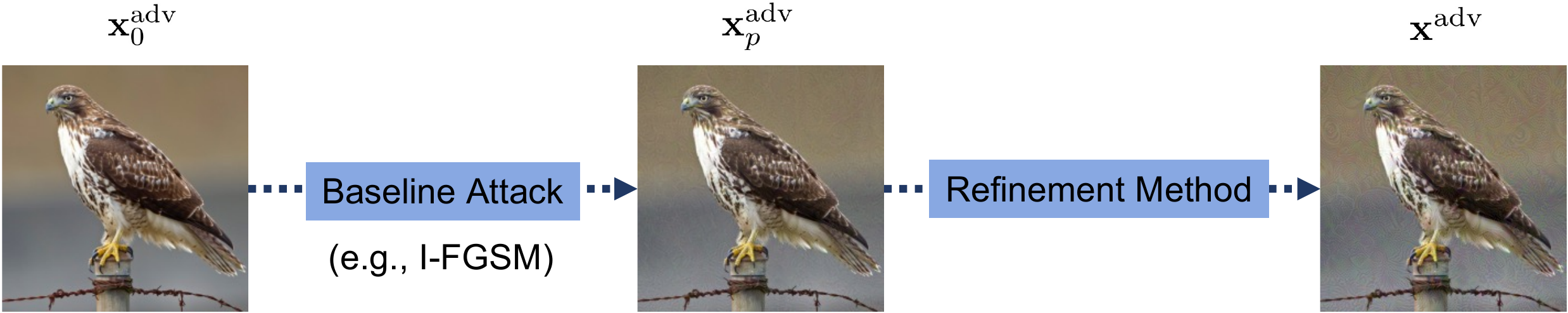}\vskip -0.1in
    \caption{An overview of the problem setting in~\cite{li2020yet},~\cite{Huang2019black}, and this paper, in which a baseline attack was performed in advance.}
    \vskip -0.05in
\label{fig:overview}
\end{figure}

The concerned problem setting throughout this paper is that, some initial adversarial examples have already been crafted using a baseline method like I-FGSM, and we aim to develop a method to refine the adversarial examples and enhance their transferability across a set of unseen victim models. See Figure~\ref{fig:overview}. 
The transferability is experimentally assessed by evaluating the attack success rates on a variety of victim models, using adversarial examples generated on a single source model~\footnote{Regardless the success rate on the source model, which is generally near 100\%}.

\begin{table*}[t!]
 \caption{Comparison of ILA++ to ILA and the baseline attack, \ie, I-FGSM. Success rate of transfer-based attacks on ImageNet is reported, with $\ell_\infty$ constraint in the untargeted setting. We use the symbol * to indicate when the source model is used as the target. Average is obtained on models that are different from the source model.}\label{tab:imagenet}
 \begin{center}\resizebox{0.999\linewidth}{!}{
   \begin{tabular}{C{1.1in}C{0.35in}C{0.5in}C{0.5in}C{0.58in}C{0.5in}C{0.5in}C{0.58in}C{0.5in}C{0.5in}C{0.5in}C{0.5in}C{0.53in}C{0.45in}}

    \toprule
    Method  & $\epsilon$ & ResNet-50* & VGG-19 \cite{Simonyan2015}  & ResNet-152$\,$\cite{He2016}  & Inception v3 \cite{Szegedy2016}  & DenseNet \cite{Huang2017densely} & MobileNet v2 \cite{Sandler2018mobilenetv2} & SENet \cite{Hu2018} & ResNeXt \cite{Xie2017aggregated}  & \ WRN\ \cite{Zagoruyko2016}  & PNASNet \cite{Liu2018}  & MNASNet \cite{Tan2019mnasnet} & Average  \\
    \midrule
    \multirow{3}{*}{I-FGSM} & 16/255 & 100.00\% & 61.50\% & 52.82\% & 30.86\% & 57.36\% & 58.92\% & 38.12\% & 48.88\% & 48.92\% & 28.92\% & 57.20\% & 48.35\% \\
 & 8/255  & 100.00\% & 38.90\% & 29.36\% & 15.36\% & 34.86\% & 37.66\% & 17.76\% & 26.30\% & 26.26\% & 13.04\% & 35.08\% & 27.46\% \\
 & 4/255  & 100.00\% & 18.86\% & 11.28\% & 6.66\%  & 15.44\% & 18.36\% & 5.72\%  & 9.58\%  & 9.98\%  & 4.14\%  & 17.02\% & 11.70\% \\
    \midrule
\multirow{3}{*}{I-FGSM+ILA} & 16/255 & 99.96\% & 93.62\% & 91.50\% & 73.10\% & 92.46\% & 91.86\% & 84.22\% & 89.94\% & 89.30\% & 77.72\% & 90.78\% & 87.45\% \\
& 8/255  & 99.96\% & 72.86\% & 66.80\% & 38.58\% & 68.18\% & 69.64\% & 50.14\% & 62.56\% & 62.46\% & 40.80\% & 68.16\% & 60.02\% \\
& 4/255  & 99.94\% & 38.90\% & 29.42\% & 13.48\% & 34.04\% & 36.22\% & 18.62\% & 26.36\% & 27.42\% & 12.82\% & 35.88\% & 27.32\% \\
    \midrule
\multirow{3}{*}{I-FGSM+RR (ILA++)} & 16/255 & 99.96\% & \textbf{94.24\%} & \textbf{92.22\%} & \textbf{75.26\%} & \textbf{93.22\%} & \textbf{92.16\%} & \textbf{85.74\%} & \textbf{90.94\%} & \textbf{90.20\%} & \textbf{79.96\%} & \textbf{91.72\%} & \textbf{88.57\%} \\
& 8/255  & 99.96\% & \textbf{75.42\%} & \textbf{70.50\%} & \textbf{41.90\%} & \textbf{71.74\%} & \textbf{72.54\%} & \textbf{53.76\%} & \textbf{66.32\%} & \textbf{66.30\%} & \textbf{44.78\%} & \textbf{70.56\%} & \textbf{63.38\%} \\
& 4/255  & 99.94\% & \textbf{42.76\%} & \textbf{33.14\%} & \textbf{15.90\%} & \textbf{38.10\%} & \textbf{39.76\%} & \textbf{21.06\%} & \textbf{30.28\%} & \textbf{31.02\%} & \textbf{15.20\%} & \textbf{38.72\%} & \textbf{30.59\%} \\
 
  \bottomrule
   \end{tabular}   
   }
 \end{center}\vskip -0.15in
\end{table*}

\subsection{Intermediate-level Attacks}\label{sec:ila}

Suppose that I-FGSM~\cite{Kurakin2017} has already been performed on a source model as a baseline attack, as has been illustrated in Figure~\ref{fig:overview}, the goal of our work is to enhance the transferability of the I-FGSM results. 

Given a benign example $\mathbf x$, the update rule of I-FGSM is: for the $t$-th step,
\begin{equation}\label{eq:ifgsm}
    \mathbf x^{\mathrm{adv}}_{t+1} = \mathbf \Pi_\Psi(\mathbf x^{\mathrm{adv}}_t + \alpha\cdot \mathrm{sgn}(\nabla f(\mathbf x^{\mathrm{adv}}_t, y))),
\end{equation}
in which $\alpha>0$ is the attack step size, $\mathbf x^{\mathrm{adv}}_0=\mathbf x$, and $\mathbf \Pi_\Psi(\cdot)$ projects its input onto the set of valid images (\ie, $\Psi$). If in total $p$ steps of the I-FGSM update were performed, we could collect $p$ temporary results $\mathbf x^{\mathrm{adv}}_0,\ldots, \mathbf x^{\mathrm{adv}}_t, \ldots, \mathbf x^{\mathrm{adv}}_{p-1}$ together with the final baseline result $\mathbf x^{\mathrm{adv}}_{p}$. We also collected the value of their corresponding prediction loss (\ie, the cross-entropy loss evaluating their prediction correctness), denoted by $l_0,\ldots, l_t,\ldots, l_p$.
The middle-layer features are represented as $\mathbf h^{\mathrm{adv}}_u=g(\mathbf x^{\mathrm{adv}}_u)$, for $u=0,
\ldots,t, \ldots, p$.

To achieve an adversarial example with improved transferability on the basis of I-FGSM~\cite{Kurakin2017}, we proposed~\cite{li2020yet} to solve
\begin{equation}\label{eq:opt0}
    \max_{\mathbf{\Delta}_{\mathbf x}}\ (g(\mathbf x+\mathbf{\Delta}_{\mathbf x}) - \mathbf h^{\mathrm{adv}}_0)^T \mathbf w^\ast, \quad \mathrm{s.t.}\ (\mathbf x+\mathbf{\Delta}_{\mathbf x}) \in \Psi. 
\end{equation}

In~\cite{li2020yet}, $\mathbf w^\ast = (\mathbf H^T \mathbf H + \lambda \mathbf I_m)^{-1} \mathbf H^T \mathbf r$ was obtained from a closed-form solution of a ridge regression problem, in which the $t$-th row of $\mathbf H\in\mathbb R^{(p+1)\times m}$ was $(\mathbf h_t^{\mathrm{adv}}-\mathbf h^{\mathrm{adv}}_0)^T$ and the $t$-th entry of $\mathbf r\in\mathbb R^{p+1}$ was $l_t$. We showed in that paper that an approximation to $\mathbf w^\ast$ could be obtained using $\mathbf w^\ast\approx\mathbf H^T \mathbf r$, in order to gain higher computational efficiency (see Appendix~\ref{sec:app} for a brief introduction of the approximation). That is, we can turn to solve
\begin{equation}\label{eq:opt0_app}
    \max_{\mathbf{\Delta}_x}\ (g(\mathbf x+\mathbf{\Delta}_\mathbf x) - \mathbf h^{\mathrm{adv}}_0)^T \mathbf H^T \mathbf r, \quad \mathrm{s.t.}\ (\mathbf x+\mathbf{\Delta}_\mathbf x) \in \Psi.
\end{equation}

In a related paper, Huang \etal~\cite{Huang2019black} also proposed to operate on the middle-layer features of the source model for gaining higher transferability, and their work inspired ours. Specifically, they proposed to solve
\begin{equation}\label{eq:opt1}
    \max_{\mathbf{\Delta}_x}\ (g(\mathbf x+\mathbf{\Delta}_\mathbf x) - \mathbf h^{\mathrm{adv}}_0)^T (\mathbf h^{\mathrm{adv}}_p - \mathbf h^{\mathrm{adv}}_0), \quad \mathrm{s.t.}\ (\mathbf x+\mathbf{\Delta}_\mathbf x) \in \Psi.
\end{equation}
Apparently, their formulation in Eq.~\eqref{eq:opt1} can be regarded as a special case of ours in Eq.~\eqref{eq:opt0_app}, in a single-step setting of I-FGSM with $p=1$. Since both Huang \etal's method and our method operate on the intermediate-level representations of DNNs, we will call them intermediate-level attack (ILA) and ILA++, respectively, in this paper.

\begin{table*}[t!]
 \caption{Comparison of different linear regression models in $\ell_\infty$ attack settings. The success rate of transfer-based attacks on ImageNet is evaluated, with different $\ell_\infty$ constraints in the untargeted setting. We use the symbol * to indicate when the source model is used as the target. Average is calculated on the models different from the source model. }\label{tab:linear_regression}
 \begin{center}\resizebox{0.999\linewidth}{!}{
   \begin{tabular}{C{1.0in}C{0.35in}C{0.5in}C{0.5in}C{0.58in}C{0.5in}C{0.5in}C{0.58in}C{0.5in}C{0.5in}C{0.5in}C{0.5in}C{0.53in}C{0.45in}}
    \toprule
    Method  & $\epsilon$ & ResNet-50* & VGG-19 \cite{Simonyan2015}  & ResNet-152$\,$\cite{He2016}  & Inception v3 \cite{Szegedy2016}  & DenseNet \cite{Huang2017densely} & MobileNet v2 \cite{Sandler2018mobilenetv2} & SENet \cite{Hu2018} & ResNeXt \cite{Xie2017aggregated}  & \ WRN\ \cite{Zagoruyko2016}  & PNASNet \cite{Liu2018}  & MNASNet \cite{Tan2019mnasnet} & Average  \\
    \midrule
    \multirow{3}{*}{I-FGSM}            & 16/255 & 100.00\% & 61.50\% & 52.82\% & 30.86\% & 57.36\% & 58.92\% & 38.12\% & 48.88\% & 48.92\% & 28.92\% & 57.20\% & 48.35\% \\
                                  & 8/255  & 100.00\% & 38.90\% & 29.36\% & 15.36\% & 34.86\% & 37.66\% & 17.76\% & 26.30\% & 26.26\% & 13.04\% & 35.08\% & 27.46\% \\
                                  & 4/255  & 100.00\% & 18.86\% & 11.28\% & 6.66\%  & 15.44\% & 18.36\% & 5.72\%  & 9.58\%  & 9.98\%  & 4.14\%  & 17.02\% & 11.70\% \\\midrule
\multirow{3}{*}{I-FGSM+RR} & 16/255 & 99.96\% & \textbf{94.24\%} & 92.22\% & 75.26\% & \textbf{93.22\%} & \textbf{92.16\%} & \textbf{85.74\%} & \textbf{90.94\%} & \textbf{90.20\%} & 79.96\% & 91.72\% & 88.57\% \\
& 8/255  & 99.96\% & 75.42\% & 70.50\% & 41.90\% & 71.74\% & 72.54\% & 53.76\% & 66.32\% & 66.30\% & 44.78\% & 70.56\% & 63.38\% \\
& 4/255  & 99.94\% & 42.76\% & 33.14\% & 15.90\% & 38.10\% & 39.76\% & 21.06\% & 30.28\% & 31.02\% & 15.20\% & 38.72\% & 30.59\% \\\midrule
\multirow{3}{*}{I-FGSM+ElasticNet} & 16/255 & 99.90\%  & 89.76\% & 84.82\% & 63.76\% & 86.48\% & 86.84\% & 76.58\% & 82.84\% & 82.72\% & 67.14\% & 86.68\% & 80.76\% \\
                                  & 8/255  & 99.92\%  & 64.62\% & 54.44\% & 29.70\% & 57.46\% & 60.30\% & 41.08\% & 51.06\% & 51.22\% & 30.88\% & 59.94\% & 50.07\% \\
                                  & 4/255  & 99.90\%  & 31.66\% & 21.64\% & 10.62\% & 26.02\% & 29.14\% & 13.58\% & 19.40\% & 20.52\% & 9.40\%  & 28.50\% & 21.05\% \\\midrule
\multirow{3}{*}{I-FGSM+SVR}        & 16/255 & 99.96\% & 94.18\% & \textbf{92.24\%} & \textbf{76.18\%} & 93.18\% & 92.06\% & 85.70\% & 90.92\% & 89.96\% & \textbf{80.02\%} & \textbf{92.10\%} & \textbf{88.65\%} \\
        & 8/255  & 99.96\% & \textbf{75.96\%} & \textbf{71.00\%} & \textbf{42.96\%} & \textbf{72.16\%} & \textbf{72.94\%} & \textbf{54.28\%} & \textbf{66.86\%} & \textbf{66.90\%} & \textbf{45.52\%} & \textbf{71.52\%} & \textbf{64.01\%} \\
        & 4/255  & 99.92\% & \textbf{43.80\%} & \textbf{34.64\%} & \textbf{16.50\%} & \textbf{38.24\%} & \textbf{41.38\%} & \textbf{21.88\%} & \textbf{31.70\%} & \textbf{32.32\%} & \textbf{15.86\%} & \textbf{39.62\%} & \textbf{31.59\%} \\\bottomrule
   \end{tabular}   
   }
 \end{center} \vskip -0.12in
\end{table*} 

\begin{figure}[ht]
	\includegraphics[width=0.49\textwidth]{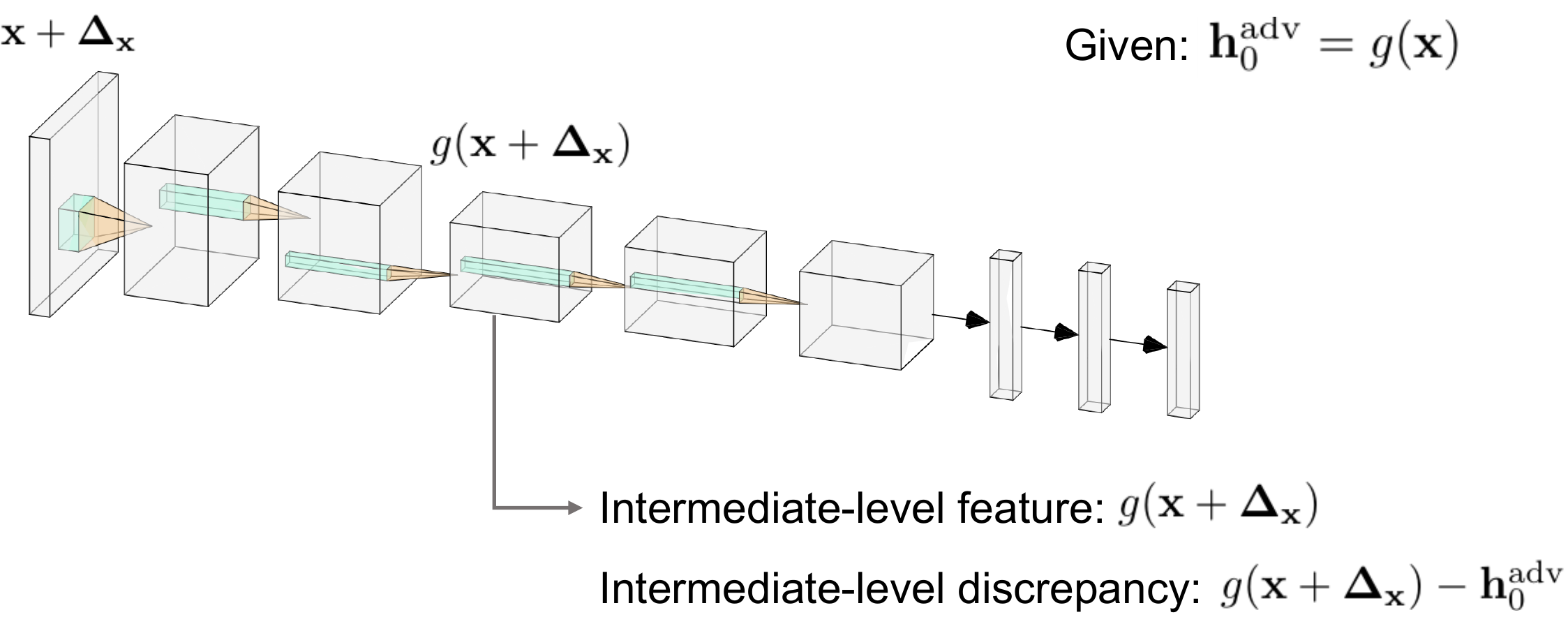}
	\caption{Illustration of how the intermediate-level discrepancy is calculated given an adversarial input $\mathbf x+\Delta_{\mathbf x}$ and its benign counterpart $\mathbf x$. We will discuss how 1) the magnitude of the intermediate-level discrepancy (\ie, $\|g(\mathbf x+\mathbf{\Delta}_\mathbf{x}) - \mathbf h^{\mathrm{adv}}_0\|$) and 2) the directional guide for optimizing $\Delta_{\mathbf x}$ would affect the obtained transferability. } 
\label{fig:pipeline}\vskip -0.15in
\end{figure}

\begin{table*}[ht]
 \caption{Comparison of different linear regression models in $\ell_2$ attack settings. The success rate of transfer-based attacks on ImageNet is evaluated, with different $\ell_2$ constraints in the untargeted setting. We use the symbol * to indicate when the source model is used as the target. Average is calculated on the models different from the source model. }\label{tab:linear_regression_l2}
 \begin{center}\resizebox{0.999\linewidth}{!}{
   \begin{tabular}{C{1.0in}C{0.35in}C{0.5in}C{0.5in}C{0.58in}C{0.5in}C{0.5in}C{0.58in}C{0.5in}C{0.5in}C{0.5in}C{0.5in}C{0.53in}C{0.45in}}
    \toprule
    Method  & $\epsilon$ & ResNet-50* & VGG-19 \cite{Simonyan2015}  & ResNet-152$\,$\cite{He2016}  & Inception v3 \cite{Szegedy2016}  & DenseNet \cite{Huang2017densely} & MobileNet v2 \cite{Sandler2018mobilenetv2} & SENet \cite{Hu2018} & ResNeXt \cite{Xie2017aggregated}  & \ WRN\ \cite{Zagoruyko2016}  & PNASNet \cite{Liu2018}  & MNASNet \cite{Tan2019mnasnet} & Average  \\
    \midrule
    \multirow{3}{*}{I-FGSM}            & 15 & 99.96\% & 78.64\% & 74.94\% & 50.02\% & 76.36\% & 75.20\% & 60.18\% & 71.16\% & 71.20\% & 51.20\% & 73.80\% & 68.27\% \\
                                  & 10 & 99.94\% & 64.48\% & 58.60\% & 33.64\% & 59.94\% & 60.92\% & 42.20\% & 53.98\% & 53.76\% & 33.94\% & 59.12\% & 52.06\% \\
                                  & 5  & 99.98\% & 36.96\% & 28.12\% & 14.68\% & 31.70\% & 34.94\% & 18.02\% & 25.40\% & 25.82\% & 12.92\% & 33.08\% & 26.16\% \\\midrule
\multirow{3}{*}{I-FGSM+RR}         & 15 & 99.92\% & 93.82\% & \textbf{91.32\%} & 75.16\% & 92.32\% & \textbf{91.74\%} & 83.28\% & \textbf{89.56\%} & 89.28\% & \textbf{78.98\%} & 91.14\% & 87.66\% \\
& 10 & 99.90\% & 84.42\% & 80.96\% & 56.72\% & 81.54\% & \textbf{82.08\%} & 66.58\% & 77.92\% & 77.26\% & 59.28\% & \textbf{81.06\%} & 74.78\% \\
& 5  & 99.90\% & 59.00\% & 51.76\% & 27.42\% & 54.06\% & 55.52\% & 35.76\% & 47.58\% & 47.82\% & 29.84\% & 53.82\% & 46.26\% \\\midrule
\multirow{3}{*}{I-FGSM+ElasticNet} & 15 & 99.92\% & 87.42\% & 82.22\% & 61.58\% & 84.42\% & 85.08\% & 71.16\% & 79.46\% & 79.34\% & 63.68\% & 84.60\% & 77.90\% \\
                                  & 10 & 99.88\% & 75.48\% & 66.82\% & 42.96\% & 69.08\% & 71.44\% & 52.42\% & 63.18\% & 63.60\% & 43.90\% & 71.14\% & 62.00\% \\
                                  & 5  & 99.74\% & 46.84\% & 38.38\% & 20.70\% & 41.66\% & 44.72\% & 26.14\% & 34.52\% & 35.22\% & 20.78\% & 43.46\% & 35.24\% \\\midrule
\multirow{3}{*}{I-FGSM+SVR}        & 15 & 99.92\% & \textbf{94.06\%} & 91.22\% & \textbf{76.16\%} & \textbf{92.38\%} & \textbf{91.74\%} & \textbf{83.48\%} & 89.42\% & \textbf{89.44\%} & 78.94\% & \textbf{91.24\%} & \textbf{87.81\%} \\
   & 10 & 99.90\% & \textbf{84.98\%} & \textbf{81.64\%} & \textbf{58.36\%} & \textbf{82.28\%} & 81.90\% & \textbf{66.74\%} & \textbf{78.40\%} & \textbf{77.88\%} & \textbf{59.94\%} & 80.96\% & \textbf{75.31\%} \\
   & 5  & 99.90\% & \textbf{59.08\%} & \textbf{53.16\%} & \textbf{29.06\%} & \textbf{55.08\%} & \textbf{56.26\%} & \textbf{36.70\%} & \textbf{48.30\%} & \textbf{48.80\%} & \textbf{30.88\%} & \textbf{55.16\%} & \textbf{47.25\%} \\
    \bottomrule
   \end{tabular}   
   }
 \end{center} \vskip -0.12in
\end{table*} 

\subsection{Linear Regression for Obtaining $h'$}

Our ILA++~\cite{li2020yet} established a direct mapping from the intermediate layer of a DNN to the prediction loss, such that how much classification loss the intermediate-level discrepancies shall lead to can be reasonably evaluated. It shows considerably better performance in comparison with the baseline attack and ILA (see Table~\ref{tab:imagenet} and other experimental results in~\cite{li2020yet}).

The effectiveness of ILA++ can be explained from different perspectives. 1) ILA++ substitutes the original nonlinear function $h$ with a linear one, and it is likely that a more linear model leads to improved transferability of adversarial examples~\cite{guo2020back}. 2) the optimization problem of ILA++ aims at maximizing the projection on a directional guide $\mathbf H^T \mathbf r$ which is a linear combination of many intermediate-level discrepancies (\ie, ``perturbations'' of features) obtained from the iterative baseline attack. Such a directional guide, \ie, $\mathbf H^T \mathbf r$, is basically a weighted sum of some baseline intermediate-level discrepancies. It leads to higher prediction loss and is thus more powerful.  

We study whether other choices of $\mathbf w^\ast$ improve adversarial transferability. 
Still, we focus on a framework in which the refined adversarial examples are obtained via solving Eq.~\eqref{eq:opt0} using off-the-shelf constrained optimization methods (\eg, FGSM~\cite{Goodfellow2015}, I-FGSM~\cite{Kurakin2017}, PGD~\cite{Madry2018}) and input gradient of $f'=h'\circ g$, rather than input gradient of $f=h\circ g$.
These optimization methods are variants of gradient ascent and they can be performed just like in their original settings, except with a different objective function, \ie, Eq.~\eqref{eq:opt0}.

We first compare linear regression models for obtaining $\mathbf w^\ast$ and learning the relationship between intermediate-level discrepancies and prediction loss.
In particular, three regression models are considered, including ridge regression (RR, also known as Tikhonov regularization) as introduced in Section~\ref{sec:ila}, ElasticNet~\cite{zou2005regularization}, and support vector regression (SVR)~\cite{platt1999probabilistic}:
\begin{equation}\label{eq:optall}
\begin{aligned}
&\mathrm{RR:}\ 
\min_{\mathbf w}\ \sum^p_{t=0} z_t^2 + \lambda \|\mathbf w\|_2^2, \\
&\mathrm{ElasticNet:}\  
\min_{\mathbf w}\ \sum^p_{t=0} z_t^2 + \lambda_1 \|\mathbf w\|_1+\lambda_2 \|\mathbf w\|^2_2, \\
&\mathrm{SVR:}\ 
\min_{\mathbf w} \frac{1}{2}\|\mathbf w\|_2^2 \quad\mathrm{s.t.}\ |z_t-b|<e, \\  
\end{aligned}
\end{equation}
in which $z_t=\mathbf w^T (\mathbf h^{\mathrm{adv}}_t-\mathbf h^{\mathrm{adv}}_0) - l_t$, and $b$ is set to zero. 

By solving different optimization problems in Eq.~\eqref{eq:optall}, we can obtain different $\mathbf w^\ast$ for Eq.~\eqref{eq:opt0}. Based on them, different refined adversarial examples can eventually be obtained by solving Eq.~\eqref{eq:opt0} then. Other sparse coding algorithms than ElasticNet~\cite{zou2005regularization} are not considered, since a sparse $\mathbf w^\ast$ in fact restricts possible directions of perturbations in the feature space, and thus lead to worse attack success rates even on the source model. We show more about this in Tables~\ref{tab:elastic_lambda_victim} and~\ref{tab:elastic_lambda_source} by varying the hyper-parameter $\lambda_1$ in Eq.~\eqref{eq:optall}.

\begin{table*}[t!]
 \caption{Evaluation of normalized $\ell_\infty$ attacks. The attack success rate on ImageNet in the untargeted setting are reported. The symbol * is used to indicate when the source model is used as the target. It can be obviously seen that the performance of these linear regression models degraded significantly in comparison with the results in Table~\ref{tab:linear_regression}. }\label{tab:normalized_linear_regression}
 \begin{center}\resizebox{0.99\linewidth}{!}{
   \begin{tabular}{C{1.0in}C{0.35in}C{0.5in}C{0.5in}C{0.58in}C{0.5in}C{0.5in}C{0.58in}C{0.5in}C{0.5in}C{0.5in}C{0.5in}C{0.53in}C{0.45in}}
    \toprule
    Method  & $\epsilon$ & ResNet-50* & VGG-19 \cite{Simonyan2015}  & ResNet-152$\,$\cite{He2016}  & Inception v3 \cite{Szegedy2016}  & DenseNet \cite{Huang2017densely} & MobileNet v2 \cite{Sandler2018mobilenetv2} & SENet \cite{Hu2018} & ResNeXt \cite{Xie2017aggregated}  & \ WRN\ \cite{Zagoruyko2016}  & PNASNet \cite{Liu2018}  & MNASNet \cite{Tan2019mnasnet} & Average  \\
    \midrule
     \multirow{3}{*}{I-FGSM}            & 16/255 & 100.00\% & \textbf{61.50\%} & \textbf{52.82\%} & \textbf{30.86\%} & \textbf{57.36\%} & \textbf{58.92\%} & \textbf{38.12\%} & \textbf{48.88\%} & \textbf{48.92\%} & \textbf{28.92\%} & \textbf{57.20\%} & \textbf{48.35\%} \\
                                     & 8/255  & 100.00\% & \textbf{38.90\%} & \textbf{29.36\%} & 15.36\% & \textbf{34.86\%} & \textbf{37.66\%} & \textbf{17.76\%} & \textbf{26.30\%} & \textbf{26.26\%} & \textbf{13.04\%} & \textbf{35.08\%} & \textbf{27.46\%} \\
                                  & 4/255  & 100.00\% & 18.86\% & 11.28\% & 6.66\%  & 15.44\% & 18.36\% & 5.72\%  & 9.58\%  & 9.98\%  & 4.14\%  & 17.02\% & 11.70\% \\\midrule
\multirow{3}{*}{\shortstack{I-FGSM+RR\\ (normalized)}}         & 16/255 & 99.94\% & 41.48\% & 34.52\% & 20.70\% & 40.66\% & 40.76\% & 20.20\% & 30.32\% & 31.74\% & 15.98\% & 38.46\% & 31.48\% \\
& 8/255  & 99.94\% & 35.24\% & 27.94\% & 15.90\% & 33.72\% & 35.32\% & 14.78\% & 24.44\% & 25.04\% & 11.50\% & 32.70\% & 25.66\% \\
& 4/255  & 99.92\% & 21.48\% & 14.76\% & 8.90\%  & 19.10\% & 21.60\% & \textbf{7.18\%}  & 12.82\% & 12.86\% & \textbf{5.68\%}  & 19.44\% & 14.38\% \\\midrule
\multirow{3}{*}{\shortstack{I-FGSM+ElasticNet\\ (normalized)}} & 16/255 & 99.78\%  & 24.36\% & 16.26\% & 11.06\% & 21.64\% & 24.94\% & 9.06\%  & 14.12\% & 15.14\% & 7.40\%  & 23.66\% & 16.76\% \\
                                  & 8/255  & 99.62\%  & 19.00\% & 11.58\% & 7.86\%  & 15.84\% & 19.46\% & 6.22\%  & 10.04\% & 10.46\% & 4.92\%  & 18.70\% & 12.41\% \\
                                  & 4/255  & 98.26\%  & 10.46\% & 5.66\%  & 4.78\%  & 8.62\%  & 12.30\% & 2.96\%  & 5.00\%  & 5.06\%  & 2.36\%  & 10.90\% & 6.81\%  \\\midrule
\multirow{3}{*}{\shortstack{I-FGSM+SVR\\ (normalized)}}        & 16/255 & 99.94\% & 39.86\% & 33.66\% & 20.38\% & 39.40\% & 39.72\% & 18.84\% & 29.72\% & 30.46\% & 15.46\% & 37.28\% & 30.48\% \\
& 8/255  & 99.94\% & 35.16\% & 27.00\% & \textbf{16.24\%} & 33.04\% & 34.32\% & 14.56\% & 23.94\% & 24.68\% & 11.26\% & 31.80\% & 25.20\% \\
& 4/255  & 99.92\% & \textbf{21.60\%} & \textbf{15.12\%} &\textbf{9.42\%}  & \textbf{19.66\%} & \textbf{22.14\%} & 7.08\%  & \textbf{13.08\%} & \textbf{13.12\%} & 5.36\%  & \textbf{20.26\%} & \textbf{14.68\%} \\\bottomrule
   \end{tabular}   
   }
 \end{center} 
\end{table*} 

\begin{table*}[t!]
 \caption{Evaluation of normalized $\ell_2$ attacks. The attack success rates on ImageNet in the untargeted setting are reported. The symbol * is used to indicate when the source model is used as the target. It can be obviously seen that the performance of these linear regression models degraded significantly in comparison with the results in Table~\ref{tab:linear_regression_l2}. }\label{tab:normalized_linear_regression_l2}
 \begin{center}\resizebox{0.99\linewidth}{!}{
   \begin{tabular}{C{1.0in}C{0.35in}C{0.5in}C{0.5in}C{0.58in}C{0.5in}C{0.5in}C{0.58in}C{0.5in}C{0.5in}C{0.5in}C{0.5in}C{0.53in}C{0.45in}}
    \toprule
    Method  & $\epsilon$ & ResNet-50* & VGG-19 \cite{Simonyan2015}  & ResNet-152$\,$\cite{He2016}  & Inception v3 \cite{Szegedy2016}  & DenseNet \cite{Huang2017densely} & MobileNet v2 \cite{Sandler2018mobilenetv2} & SENet \cite{Hu2018} & ResNeXt \cite{Xie2017aggregated}  & \ WRN\ \cite{Zagoruyko2016}  & PNASNet \cite{Liu2018}  & MNASNet \cite{Tan2019mnasnet} & Average  \\
    \midrule
\multirow{3}{*}{I-FGSM}            & 15 & 99.96\% & \textbf{78.64\%} & \textbf{74.94\%} & \textbf{50.02\%} & \textbf{76.36\%} & \textbf{75.20\%} & \textbf{60.18\%} & \textbf{71.16\%} & \textbf{71.20\%} & \textbf{51.20\%} & \textbf{73.80\%} & \textbf{68.27\%} \\
   & 10 & 99.94\% & \textbf{64.48\%} & \textbf{58.60\%} & 33.64\% & \textbf{59.94\%} & \textbf{60.92\%} & \textbf{42.20\%} & \textbf{53.98\%} & \textbf{53.76\%} & \textbf{33.94\%} & \textbf{59.12\%} & \textbf{52.06\%} \\
& 5  & 99.98\% & 36.96\% & 28.12\% & 14.68\% & 31.70\% & 34.94\% & 18.02\% & 25.40\% & 25.82\% & 12.92\% & 33.08\% & 26.16\% \\\midrule
\multirow{3}{*}{\shortstack{I-FGSM+RR\\ (normalized)}}         & 15 & 99.92\% & 72.46\% & 70.86\% & 49.46\% & 73.32\% & 71.30\% & 51.44\% & 65.04\% & 66.22\% & 44.96\% & 70.30\% & 63.54\% \\
& 10 & 99.90\% & 59.70\% & 57.20\% & \textbf{36.76\%} & 60.46\% & 59.06\% & 37.92\% & 51.22\% & 51.52\% & 33.30\% & 57.94\% & 50.51\% \\
& 5  & 99.92\% & \textbf{39.52\%} & \textbf{33.10\%} & 19.70\% & \textbf{38.20\%} & \textbf{39.08\%} & \textbf{19.28\%} & \textbf{30.08\%} & \textbf{30.04\%} & \textbf{15.92\%} & \textbf{37.14\%} & \textbf{30.21\%} \\\midrule
\multirow{3}{*}{\shortstack{I-FGSM+ElasticNet\\ (normalized)}} & 15 & 99.88\% & 51.54\% & 42.72\% & 29.66\% & 47.20\% & 49.94\% & 30.50\% & 38.88\% & 39.46\% & 25.06\% & 50.44\% & 40.54\% \\
& 10 & 99.84\% & 39.44\% & 31.16\% & 20.30\% & 35.42\% & 38.80\% & 20.96\% & 28.82\% & 29.32\% & 17.68\% & 39.48\% & 30.14\% \\
& 5  & 98.86\% & 22.96\% & 17.10\% & 10.92\% & 20.08\% & 23.82\% & 9.84\%  & 15.18\% & 14.76\% & 8.26\%  & 22.42\% & 16.53\% \\\midrule
\multirow{3}{*}{\shortstack{I-FGSM+SVR\\ (normalized)}}        & 15 & 99.92\% & 70.08\% & 68.86\% & 48.80\% & 70.86\% & 69.14\% & 49.80\% & 62.46\% & 63.62\% & 43.66\% & 68.24\% & 61.55\% \\
& 10 & 99.90\% & 58.10\% & 55.12\% & 36.46\% & 58.60\% & 57.76\% & 35.96\% & 49.74\% & 49.74\% & 32.10\% & 56.40\% & 49.00\% \\
& 5  & 99.92\% & 38.70\% & 32.30\% & \textbf{19.76\%} & 37.44\% & 38.36\% & 19.10\% & 29.80\% & 29.54\% & 15.44\% & 36.40\% & 29.68\% \\\bottomrule
   \end{tabular}   
   }
 \end{center} \vskip -0.16in
\end{table*} 

\begin{table}[ht!]
 \caption{Average success rate of I-FGSM+ElasticNet on models \emph{excluding ResNet-50} in the untargeted $\ell_\infty$ setting. }\label{tab:elastic_lambda_victim}
 \begin{center}\resizebox{0.999\linewidth}{!}{
   \begin{tabular}{C{1.0in}C{0.38in}C{0.55in}C{0.55in}C{0.55in}C{0.55in}}
    \toprule
    Method  & $\epsilon$ &$\lambda_1=0.05$ & $\lambda_1=0.1$ & $\lambda_1=0.5$ & $\lambda_1=1.0$\\
    \midrule
    \multirow{3}{*}{I-FGSM+ElasticNet} & 16/255 & 80.76\% & 78.06\% & 69.69\% & 65.24\%  \\
 & 8/255 & 50.07\% & 46.45\% & 36.40\% & 31.97\%  \\
 & 4/255 & 21.05\% & 19.31\% & 14.04\% & 11.87\% \\ \bottomrule
   \end{tabular}   
   }
 \end{center} \vskip -0.12in
\end{table} 

\begin{table}[ht!]
 \caption{Success rate of I-FGSM+ElasticNet on the source model, \emph{\ie, ResNet-50}, in the untargeted $\ell_\infty$ setting. }\label{tab:elastic_lambda_source}
 \begin{center}\resizebox{0.999\linewidth}{!}{
   \begin{tabular}{C{1.0in}C{0.38in}C{0.55in}C{0.55in}C{0.55in}C{0.55in}}
    \toprule
    Method  & $\epsilon$ &$\lambda_1=0.05$ & $\lambda_1=0.1$ & $\lambda_1=0.5$ & $\lambda_1=1.0$\\
    \midrule
    \multirow{3}{*}{I-FGSM+ElasticNet} & 16/255  & 99.90\% & 99.92\% & 99.78\% & 99.62\%  \\
 & 8/255  & 99.92\% & 99.90\% & 99.36\% & 98.16\%  \\
 & 4/255  & 99.90\% & 99.80\% & 97.48\% & 92.92\% \\ \bottomrule
   \end{tabular}   
   }
 \end{center} 
\end{table}

\textbf{Setup. } An experiment on ImageNet was setup to compare the three options in Eq.~\eqref{eq:optall}. Similar to other experiments, we used a pre-trained ResNet-50~\cite{He2016} as the source model and evaluated attack success rates of the generated adversarial examples on a variety of victim models. 
In our framework, a baseline attack I-FGSM was performed in advance, and then the obtained baseline result was enhanced by solving Eq.~\eqref{eq:opt0} using off-the-shelf methods (here, I-FGSM again), in which $\mathbf w^\ast$ was obtained from any of the regression models.
The step size of I-FGSM is set to be $1/255$ and $\epsilon/5$ for $\ell_\infty$ and $\ell_2$ attacks, respectively. 
We run $100$ iterations for I-FGSM and we let $p=10$ for attacks using temporary results from the baseline attack.
See Section~\ref{sec:exp} for details.

\textbf{Results.} Tables~\ref{tab:linear_regression} and~\ref{tab:linear_regression_l2} summarize the attack success rates using different linear regression models, in $\ell_\infty$ and $\ell_2$ settings, respectively. 
It can be seen that all the three linear regression methods led to improved attack transferability.  
Tables~\ref{tab:elastic_lambda_victim} and~\ref{tab:elastic_lambda_source} report the performance of I-FGSM+ElasticNet with $\lambda_1$ ranging from $0.05$ to $1.0$, and, in Tables~\ref{tab:linear_regression} and~\ref{tab:linear_regression_l2}, we provide detailed results with $\lambda_1=0.05$.
Apparently, smaller $\lambda_1$ leads to higher success rates using I-FGSM+ElasticNet, yet we did not test with further smaller $\lambda_1$, since solution to the ElasticNet optimization became inaccurate, as suggested by the Python \texttt{sklearn} package.
As $\lambda_1\rightarrow 0$, ElasticNet boils down to ridge regression, and we shall observe similar performance on I-FGSM+RR and I-FGSM+ElasticNet when $\lambda_1$ approaches $0$. 
$\lambda$ and $C$ were set to $1e10$ and $1e-10$, respectively, for obtaining the I-FGSM+RR and I-FGSM+SVR results in Table~\ref{tab:linear_regression} and~\ref{tab:linear_regression_l2}. Note that $C$ is inversely proportional to the strength of the SVR regularization, and $\lambda$ regularizes RR. How $\lambda$ and $C$ affect the performance of I-FGSM+RR and I-FGSM+SVR will be discussed in the following subsection.


\subsection{Maximizing Feature Distortion}\label{sec:magnitude}

Solving the optimization problem in Eq.~\eqref{eq:opt0} indicates that: 1) maximizing the middle-layer distortion (\ie, the magnitude of the intermediate-level feature discrepancy $g(\mathbf x+\Delta_{\mathbf x})-\mathbf h^{\mathrm{adv}}_0$) invoked by the adversarial example $\mathbf x+\Delta_{\mathbf x}$, and 2) maximizing the cosine similarity between $g(\mathbf x+\Delta_{\mathbf x})-\mathbf h^{\mathrm{adv}}_0$ and a reasonably chosen directional guide (\ie, $\mathbf w^\ast/\|\mathbf w^\ast\|$). It is of yet unclear whether both the two factors are essential for improving the transferability. In this subsection and the following subsection, we aim to shed more light on it.
We will first show how the magnitude of the intermediate-level discrepancy (\ie, $\|g(\mathbf x+\Delta_{\mathbf x})-\mathbf h^{\mathrm{adv}}_0\|$) affects transferability of the refined adversarial examples and attack success rates.

\begin{figure}[]
	\includegraphics[width=0.48\textwidth]{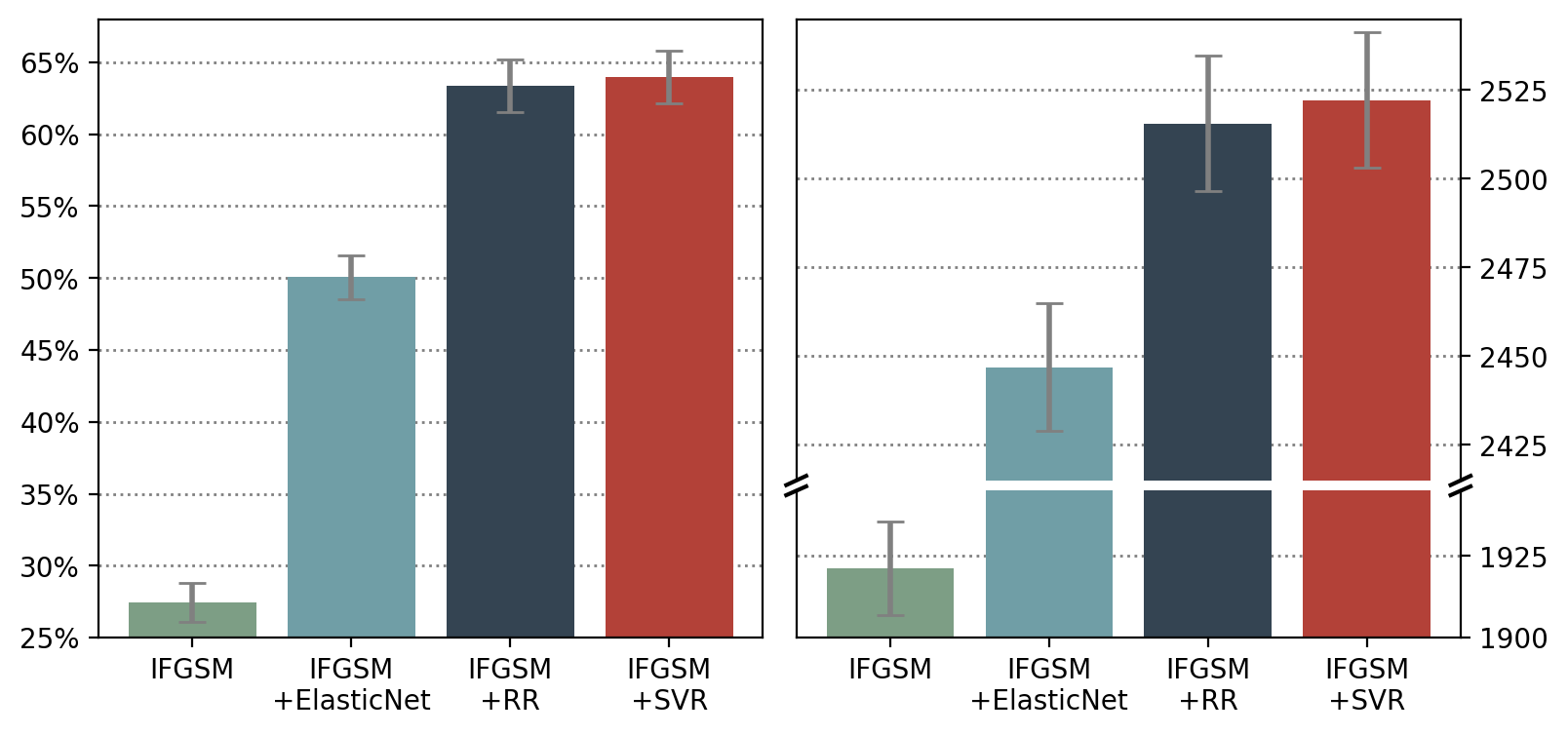}\vskip -0.1in
	\caption{On the left is the average success rate of the four methods, and on the right the average $\ell_2$ magnitude of the intermediate-level discrepancies. Apparently, they are positively correlated, and larger intermediate-level distortions lead to higher attack success rates on the victim models. } 
\label{fig:suc_mag}\vskip -0.2in
\end{figure}

We show in Figure~\ref{fig:suc_mag} the magnitude of intermediate-level discrepancies obtained using the four methods (\ie, I-FGSM, I-FGSM+RR, I-FGSM+ElasticNet, and I-FGSM+SVR), together with the obtained average attack success rates.
The error bar in each subplot has been uniformly scaled to make the illustration clearer, considering that both the magnitude and the attack success rate vary drastically across victim models.
As can be observed, the magnitude of intermediate-level discrepancies is positively correlated with the average success rate and the transferability. 

\begin{figure}[ht]
\begin{center}
\subfloat[$\lambda$ in I-FGSM+RR]{\label{fig:3a}
\includegraphics[width=0.46\textwidth]{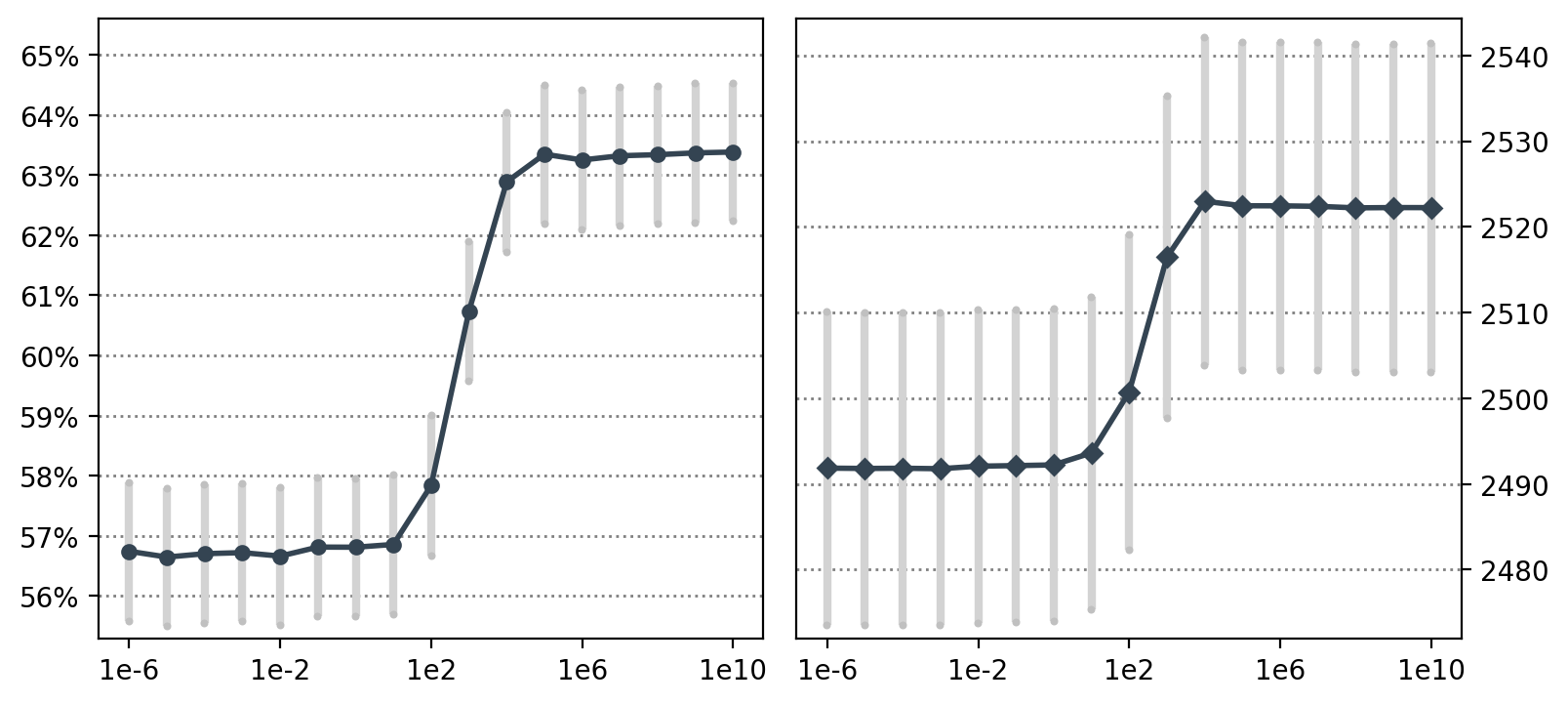}}\hskip 18pt
\subfloat[$C$ in I-FGSM+SVR]{\label{fig:3b}
\includegraphics[width=0.46\textwidth]{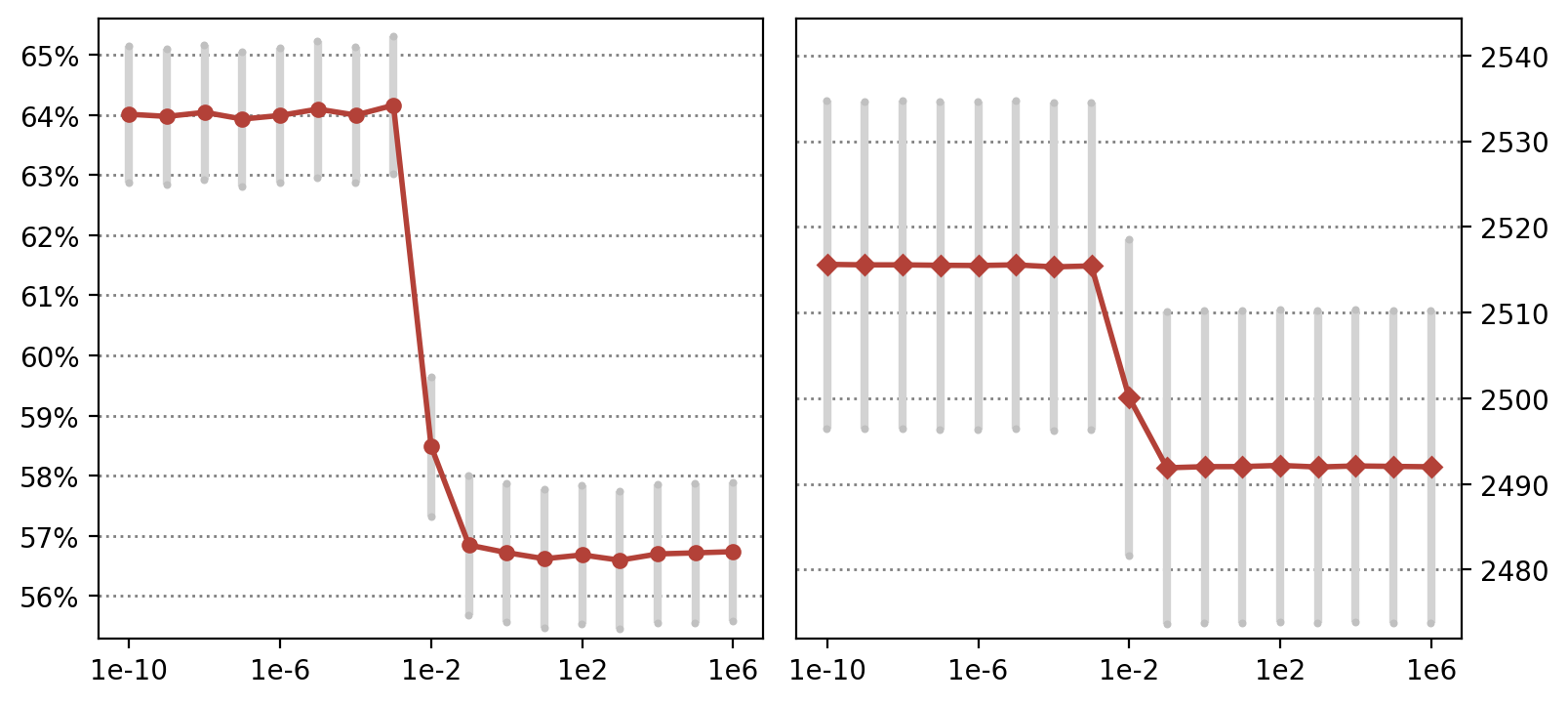}}\hskip 18pt
\caption{How the hyper-parameters of the linear regression methods affect the average success rate and middle-layer distortion (\ie, the magnitude of intermediate-level discrepancies). On the left subplots, we report the success rates, and on the right subplots, we report the magnitudes. $\epsilon=8/255$.}
\label{fig:hyper-parameters-all}
\end{center}
\vskip -0.2in
\end{figure}

Some of the hyper-parameters in fact make considerable impact on the intermediate-level distortions, \eg, $\lambda$ in the formulation of ridge regression, $\lambda_1$ and $\lambda_2$ in ElasticNet, and $C$ in SVR.
We have discussed $\lambda_1$ in Tables~\ref{tab:elastic_lambda_victim} and~\ref{tab:elastic_lambda_source}, and we know the other regularization hyper-parameter (\ie, $\lambda_2$) in ElasticNet plays the same role as $\lambda$ in ridge regression.
Therefore, here we mostly report the results of I-FGSM+RR and I-FGSM+SVR by varying $\lambda$ and $C$, respectively.
We can observe in Figure~\ref{fig:hyper-parameters-all} that, when larger intermediate-level distortions were obtained (by tuning the hyper-parameters), higher attack success rates were obtained, which also shows the positive correlation between the transferability of adversarial examples and the magnitude of the intermediate-level discrepancies.
Yet, Figure~\ref{fig:hyper-parameters-all} also shows that sensitivity of the attack performance to the hyper-parameters ($\lambda$ and $C$) only exists in a limited region, indicating that the power of $\mathbf w^\ast$ for achieving large distortion and high transferability does not always change consistently with $\lambda$ or $C$.

We further test how these methods perform when the intermediate-level feature distortion is not encouraged to be enlarged in linear regression.
We consider linear mappings between the \emph{normalized} intermediate-level discrepancies and the prediction loss of the source model. 
With such a mapping (denoted by $\tilde{\mathbf w}^\ast$ here), we can rewrite the optimization problem as:
\begin{equation}\label{eq:opt2}
    \max_{\mathbf{\Delta}_{\mathbf x}}\ \left(\frac{g(\mathbf x+\mathbf{\Delta}_{\mathbf x})- \mathbf h^{\mathrm{adv}}_0}{\|g(\mathbf x+\mathbf{\Delta}_{\mathbf x})- \mathbf h^{\mathrm{adv}}_0\|_2} \right)^T \tilde{\mathbf w}^\ast, \  \mathrm{s.t.}\ (\mathbf x+\mathbf{\Delta}_{\mathbf x}) \in \Psi
\end{equation}
in which the magnitude of $g(\mathbf x+\mathbf{\Delta}_{\mathbf x}) - \mathbf h^{\mathrm{adv}}_0$ is not encouraged to be maximized in obtaining the final refined adversarial example. 

For solving the optimization problem in Eq.~\eqref{eq:opt2}, attacks can similarly be performed, just like for solving Eq.~\eqref{eq:opt0}.
Tables~\ref{tab:normalized_linear_regression} and~\ref{tab:normalized_linear_regression_l2} demonstrate that such 
``normalized'' attacks fail to achieve satisfactory results on the ImageNet victim models. In the $\ell_\infty$ settings, I-FGSM+RR shows an average success rate of $29.17\%$, $22.63\%$, and $11.39\%$, under $\epsilon=16/255$, $8/255$, and $4/255$, respectively. The obtained attack success rates are generally no higher than the baseline I-FGSM results under  $\epsilon=16/255$ and $8/255$ and far lower than those obtained when no normalization is enforced. Similar observations can be made on the $\ell_2$ attack settings. We thus know from the experimental results that the magnitude of the intermediate-level discrepancies is an essential factor. 

\begin{table*}[t!]
 \caption{Evaluation of random directional guides in the input space. The attack success rates on ImageNet are reported, with $\ell_\infty$ constraints in the untargeted setting. The symbol * is used to indicate when the source model is used as the target. We see that random directional guides lead to poor transferability in comparison to the results in Table~\ref{tab:linear_regression}.}\label{tab:random_input}
 \begin{center}\resizebox{0.99\linewidth}{!}{
   \begin{tabular}{C{0.85in}C{0.35in}C{0.5in}C{0.5in}C{0.58in}C{0.5in}C{0.5in}C{0.58in}C{0.5in}C{0.5in}C{0.5in}C{0.5in}C{0.53in}C{0.45in}}
    \toprule
    Method  & $\epsilon$ & ResNet-50* & VGG-19 \cite{Simonyan2015}  & ResNet-152$\,$\cite{He2016}  & Inception v3 \cite{Szegedy2016}  & DenseNet \cite{Huang2017densely} & MobileNet v2 \cite{Sandler2018mobilenetv2} & SENet \cite{Hu2018} & ResNeXt \cite{Xie2017aggregated}  & \ WRN\ \cite{Zagoruyko2016}  & PNASNet \cite{Liu2018}  & MNASNet \cite{Tan2019mnasnet} & Average  \\
    \midrule
\multirow{3}{*}{I-FGSM}            & 16/255 & 100.00\% & 61.50\% & \textbf{52.82\%} & 30.86\% & 57.36\% & 58.92\% & 38.12\% & \textbf{48.88\%} & 48.92\% & \textbf{28.92\%} & 57.20\% & 48.35\% \\
& 8/255  & 100.00\% & \textbf{38.90\%} & \textbf{29.36\%} & \textbf{15.36\%} & \textbf{34.86\%} & \textbf{37.66\%} & \textbf{17.76\%} & \textbf{26.30\%} & \textbf{26.26\%} & \textbf{13.04\%} & \textbf{35.08\%} & \textbf{27.46\%} \\
   & 4/255  & 100.00\% & \textbf{18.86\%} & \textbf{11.28\%} &\textbf{ 6.66\%}  & \textbf{15.44\%} & \textbf{18.36\%} &\textbf{ 5.72\%}  &\textbf{ 9.58\%}  &\textbf{ 9.98\%}  &\textbf{ 4.14\%}  & \textbf{17.02\%} & \textbf{11.70\%} \\\midrule
\multirow{3}{*}{Rand+RR}         & 16/255 & 97.82\% & 66.26\% & 50.80\% & 31.44\% & 61.50\% & \textbf{76.50\%} & \textbf{41.60\%} & 45.50\% & \textbf{50.50\%} & 17.74\% & 69.56\% & 51.14\% \\
& 8/255  & 90.94\% & 35.66\% & 18.96\% & 9.96\%  & 27.94\% & 36.06\% & 12.98\% & 16.32\% & 19.86\% & 5.18\%  & 29.58\% & 21.25\% \\
& 4/255  & 62.86\% & 12.72\% & 5.30\%  & 4.20\%  & 9.22\%  & 11.20\% & 3.52\%  & 4.64\%  & 5.98\%  & 1.54\%  & 9.16\%  & 6.75\%  \\\midrule
\multirow{3}{*}{Rand+ElasticNet}         & 16/255 & 98.56\% & 64.14\% & 44.60\% & 21.86\% & 47.30\% & 64.02\% & 40.10\% & 43.44\% & 48.34\% & 19.08\% & 63.20\% & 45.61\% \\
& 8/255  & 93.22\% & 26.58\% & 13.26\% & 7.78\%  & 17.00\% & 24.86\% & 9.96\%  & 10.90\% & 14.30\% & 4.12\%  & 23.56\% & 15.23\% \\
& 4/255  & 59.42\% & 7.84\%  & 2.96\%  & 3.84\%  & 5.34\%  & 8.60\%  & 2.06\%  & 2.62\%  & 3.28\%  & 1.18\%  & 7.78\%  & 4.55\%  \\\midrule
\multirow{3}{*}{Rand+SVR}         & 16/255 & 97.76\% & \textbf{66.52\%} & 50.84\% & \textbf{31.68\%} & \textbf{61.82\%} & 76.04\% & 41.54\% & 45.84\% & 50.30\% & 17.66\% & \textbf{69.74\%} & \textbf{51.20\%} \\
& 8/255  & 91.02\% & 35.32\% & 18.92\% & 10.06\% & 27.84\% & 36.06\% & 12.86\% & 16.52\% & 19.74\% & 5.40\%  & 29.44\% & 21.22\% \\
& 4/255  & 62.86\% & 12.62\% & 4.92\%  & 4.46\%  & 9.24\%  & 11.28\% & 3.24\%  & 4.36\%  & 5.74\%  & 1.52\%  & 8.80\%  & 6.62\%  \\\bottomrule
   \end{tabular}   
   }
 \end{center} 
\end{table*} 

\begin{table*}[t!]
 \caption{Evaluation of random directional guides generated directly in the feature space (indicated as Rand$^\dagger$). The attack success rates on ImageNet are reported, with $\ell_\infty$ constraints in the untargeted setting. The symbol * is used to indicate when the source model is used as the target. We see that random directional guides lead to poor transferability. }\label{tab:random_feature}
 \begin{center}\resizebox{0.99\linewidth}{!}{
   \begin{tabular}{C{0.9in}C{0.35in}C{0.5in}C{0.5in}C{0.58in}C{0.5in}C{0.5in}C{0.58in}C{0.5in}C{0.5in}C{0.5in}C{0.5in}C{0.53in}C{0.45in}}
    \toprule
    Method  & $\epsilon$ & ResNet-50* & VGG-19 \cite{Simonyan2015}  & ResNet-152$\,$\cite{He2016}  & Inception v3 \cite{Szegedy2016}  & DenseNet \cite{Huang2017densely} & MobileNet v2 \cite{Sandler2018mobilenetv2} & SENet \cite{Hu2018} & ResNeXt \cite{Xie2017aggregated}  & \ WRN\ \cite{Zagoruyko2016}  & PNASNet \cite{Liu2018}  & MNASNet \cite{Tan2019mnasnet} & Average  \\
    \midrule
     \multirow{3}{*}{I-FGSM}            & 16/255 & 100.00\% & \textbf{61.50\%} & \textbf{52.82\%} & \textbf{30.86\%} & \textbf{57.36\%} & \textbf{58.92\%} & \textbf{38.12\%} & \textbf{48.88\%} & \textbf{48.92\%} & \textbf{28.92\%} & \textbf{57.20\%} & \textbf{48.35\%} \\
                                     & 8/255  & 100.00\% & \textbf{38.90\%} & \textbf{29.36\%} & \textbf{15.36\%} & \textbf{34.86\%} & \textbf{37.66\%} & \textbf{17.76\%} & \textbf{26.30\%} & \textbf{26.26\%} & \textbf{13.04\%} & \textbf{35.08\%} & \textbf{27.46\%} \\
                                     & 4/255  & 100.00\% & \textbf{18.86\%} & \textbf{11.28\%} &\textbf{ 6.66\%}  & \textbf{15.44\%} & \textbf{18.36\%} &\textbf{ 5.72\%}  &\textbf{ 9.58\%}  &\textbf{ 9.98\%}  &\textbf{ 4.14\%}  & \textbf{17.02\%} & \textbf{11.70\%} \\\midrule
\multirow{3}{*}{Rand$^\dagger$+RR}         & 16/255 & 92.58\% & 53.46\% & 31.72\% & 22.52\% & 35.48\% & 48.10\% & 30.34\% & 30.00\% & 35.20\% & 15.74\% & 47.16\% & 34.97\% \\
& 8/255  & 68.44\% & 21.24\% & 9.46\%  & 7.84\%  & 12.64\% & 20.20\% & 8.52\%  & 9.02\%  & 10.48\% & 3.76\%  & 19.48\% & 12.26\% \\
& 4/255  & 24.40\% & 7.04\%  & 2.40\%  & 3.32\%  & 4.38\%  & 7.90\%  & 2.22\%  & 2.70\%  & 3.16\%  & 0.98\%  & 7.40\%  & 4.15\%  \\\midrule
\multirow{3}{*}{Rand$^\dagger$+ElasticNet} & 16/255 & 57.28\% & 38.56\% & 29.88\% & 19.78\% & 31.06\% & 39.40\% & 27.12\% & 28.70\% & 29.90\% & 15.06\% & 38.36\% & 29.78\% \\
& 8/255  & 45.14\% & 18.42\% & 10.44\% & 7.52\%  & 12.08\% & 18.46\% & 8.20\%  & 9.26\%  & 10.22\% & 3.66\%  & 17.72\% & 11.60\% \\
& 4/255  & 24.78\% & 6.64\%  & 2.60\%  & 3.50\%  & 4.70\%  & 8.06\%  & 1.88\%  & 2.52\%  & 2.76\%  & 1.22\%  & 7.40\%  & 4.13\%  \\\midrule
\multirow{3}{*}{Rand$^\dagger$+SVR}        & 16/255 & 92.58\% & 53.40\% & 31.64\% & 22.54\% & 35.24\% & 47.54\% & 30.78\% & 30.24\% & 35.26\% & 15.86\% & 47.08\% & 34.96\% \\
& 8/255  & 68.52\% & 20.56\% & 9.52\%  & 7.38\%  & 11.90\% & 20.16\% & 8.78\%  & 8.98\%  & 10.78\% & 3.64\%  & 19.52\% & 12.12\% \\
& 4/255  & 24.64\% & 6.96\%  & 2.46\%  & 3.26\%  & 4.26\%  & 8.16\%  & 2.12\%  & 2.78\%  & 2.92\%  & 1.20\%  & 7.46\%  & 4.16\%  \\\bottomrule
   \end{tabular}   
   }
 \end{center} \vskip -0.1in
\end{table*} 

\subsection{Direction Matters As Well}\label{sec:direction}

In Section~\ref{sec:magnitude}, we have demonstrated that the magnitude of intermediate-level discrepancies matters in improving the transferability of adversarial examples. In this subsection, we will show that the choice of directional guide also affects the final attack success rates.

We have shown several different options for obtaining $\mathbf w^\ast$. One can see that, by tuning $C$, I-FGSM+SVR outperforms I-FGSM+RR, as demonstrated in Table~\ref{tab:linear_regression} and Figure~\ref{fig:hyper-parameters-all}.
We can also observe that I-FGSM+RR finally did not lead to larger intermediate-level discrepancies than I-FGSM+SVR, implying that the choice of the directional guide (or say the direction of the feature perturbation) also matters. 

We further try using random and less adversarial vectors as directional guides. 
We shall stick with the objective function in~\eqref{eq:opt0}, except that now we are free to use any random method for obtaining $\mathbf w^\ast$. 
Several different ways can be tested, to achieve the goal. First, we can use random perturbations (rather than the baseline adversarial perturbations) to get a set of inputs. That is, we can add $\Delta_0, \ldots, \Delta_{p-1}$, in which $\Delta_t\sim \mathcal N(0, \Sigma)$, $\forall 0\leq t\leq p-1$, to $\mathbf x$ and get $p$ inputs $\mathbf x+\Delta_0, \ldots, \mathbf x+\Delta_{p-1}$. Given these inputs, the intermediate-level discrepancies and prediction loss on the source model can then be computed, and we can further obtain a linear regression model using for example RR, ElasticNet, and SVR.
An alternative method is to incorporate randomness on the hidden layer. That is, we may first compute $\mathbf h^{\mathrm adv}_0 = g(\mathbf x)$ and then add random perturbations $\Delta'_0, \ldots, \Delta'_{p-1}$, in which $\Delta'_t\sim \mathcal N(0, \Sigma')$, $\forall 0\leq t\leq p-1$, to $\mathbf h^{\mathrm adv}_0$ and evaluate the prediction loss on the basis of the perturbed features. The linear regression model is then similarly obtained and Eq.~\eqref{eq:opt0} is to be solved with such random $\mathbf w^\ast$. 

Tables~\ref{tab:random_input} and~\ref{tab:random_feature} summarize the attack success rates using the two sorts of random directions, respectively. It can be seen that, since the random directions did not lead to high prediction error even on the ResNet-50 source model, the final attack success rates on the concerned victim models are also unsatisfactory. We would like to further mention that the magnitude of intermediate-level discrepancies are, nevertheless, high. More specifically, for random directional guides, average magnitude of the obtained intermediate-level discrepancies can achieve $2077.92$, yet the final average success rate on the victim models is at most $21.25\%$ under $\epsilon=8/255$, which is even worse than the I-FGSM baseline (\ie, $27.46\%$ with $1921.30$). That is, given random (and thus non-disruptive) directional guides, it is challenging to learn adversarial examples that could achieve satisfactory attack success rates on the source model even with more significant intermediate-level perturbations. This can be ascribed to the high dimensionality of the intermediate-level feature space, and we therefore conclude that the direction matters as well.

Comparing the two sorts of random directional guides, we also conclude that random directions in the input space lead to slightly more disruptive attacks in contrast to that derived from random vectors in the feature space. We believe this is unsurprising, as random vectors in the feature space do not necessarily lie on the image manifold and can be misleading for guiding intermediate-level attacks.

\section{To Further Boost The Performance}\label{sec:diverse}

\begin{figure}[ht]
\begin{center}
\includegraphics[width=\columnwidth]{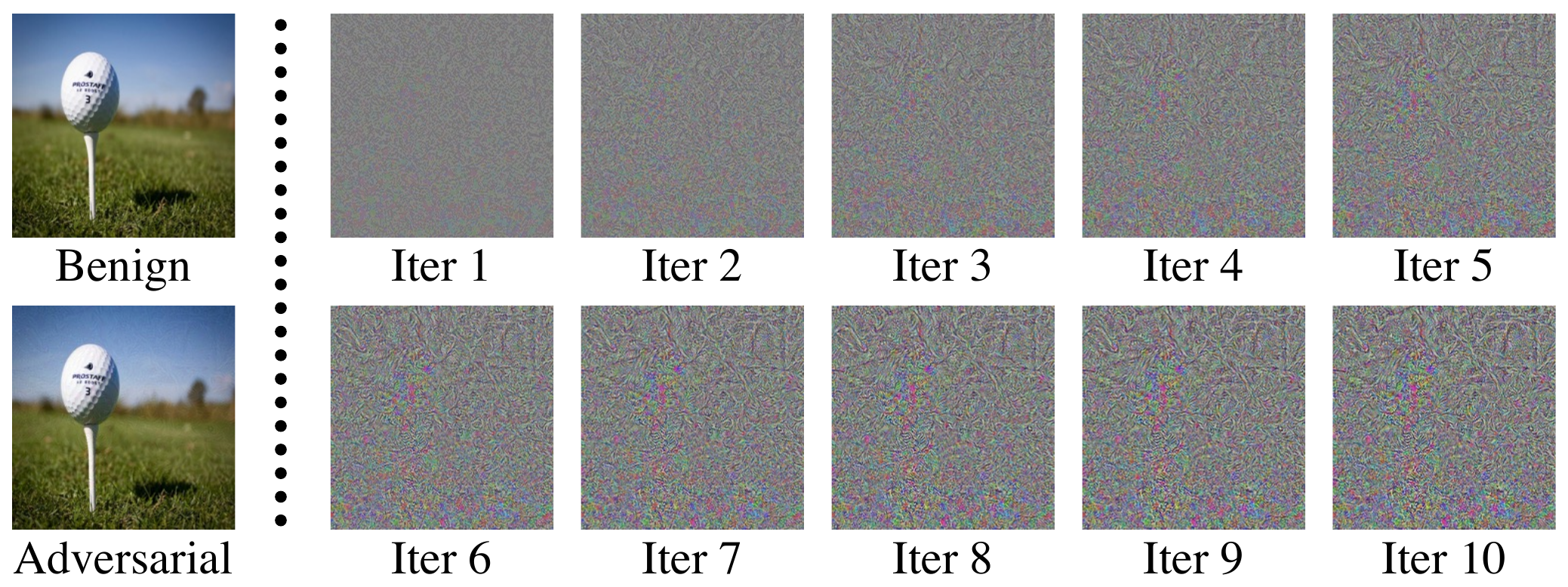}
\caption{Visualization of the obtained adversarial perturbations using I-FGSM. It can be seen that perturbations at later iterations are all very similar. $\epsilon=8/255$. }
\label{fig:vis_perturb}
\end{center}
\vskip -0.15in
\end{figure}

We conclude from Section~\ref{sec:linear} that: 1) more transferable adversarial examples can be obtained by taking advantage of some linear regression models learned from the temporary results $\mathbf x^{\mathrm{adv}}_0,\ldots, \mathbf x^{\mathrm{adv}}_t, \ldots, \mathbf x^{\mathrm{adv}}_{p-1}$ of a multi-step baseline attack, and 2) it is challenging to learn a disruptive adversarial example given only random intermediate-level feature discrepancies and their corresponding prediction loss. Nevertheless, for a $p$-step baseline attack, the method is only capable of collecting $p+1$ samples to train a linear regression model, let alone later results (with relatively large $t$) of the baseline attack are all very similar and thus less informative, as can be seen in Figure~\ref{fig:vis_perturb}. 
As such, one plausible way of further improving performance in our framework is to apply more than one baseline search and collect more diverse directional guides, if computational complexity of the baseline attack is not a primary concern. Once $\mathbf w^\ast$ is obtained, the objective in Eq.~\eqref{eq:opt0} is still utilized to encourage maximal intermediate-level distortion with these directional guides.

\begin{table*}[t!]
 \caption{Further performance gain can be achieved by performing more than one run of a baseline method and using a more powerful baseline attack, \eg, LinBP. The attack success rates on ImageNet are reported, with $\ell_\infty$ constraints in the untargeted setting. The symbol * is used to indicate when the source model is used as the target. }\label{tab:linbp_linf}
 \begin{center}\resizebox{0.99\linewidth}{!}{
   \begin{tabular}{C{1.25in}C{0.35in}C{0.5in}C{0.5in}C{0.58in}C{0.5in}C{0.5in}C{0.58in}C{0.5in}C{0.5in}C{0.5in}C{0.5in}C{0.53in}C{0.45in}}
    \toprule
    Method  & $\epsilon$ & ResNet-50* & VGG-19 \cite{Simonyan2015}  & ResNet-152$\,$\cite{He2016}  & Inception v3 \cite{Szegedy2016}  & DenseNet \cite{Huang2017densely} & MobileNet v2 \cite{Sandler2018mobilenetv2} & SENet \cite{Hu2018} & ResNeXt \cite{Xie2017aggregated}  & \ WRN\ \cite{Zagoruyko2016}  & PNASNet \cite{Liu2018}  & MNASNet \cite{Tan2019mnasnet} & Average  \\
    \midrule
\multirow{3}{*}{LinBP}         & 16/255 & 100.00\% & 93.04\% & 89.42\% & 60.72\% & 90.34\% & 88.90\% & 77.26\% & 84.38\% & 85.06\% & 63.78\% & 87.78\% & 82.07\% \\
& 8/255 & 100.00\% & 71.72\% & 58.58\% & 29.54\% & 63.42\% & 64.42\% & 41.36\% & 51.22\% & 54.62\% & 29.98\% & 62.52\% & 52.74\% \\
& 4/255 & 99.98\%  & 36.28\% & 22.48\% & 10.60\% & 28.72\% & 31.80\% & 13.40\% & 18.12\% & 20.26\% & 8.74\%  & 30.06\% & 22.05\% \\\midrule
\multirow{3}{*}{LinBP ($\times10$) +RR}         & 16/255 & 100.00\% & \textbf{97.80\%} & \textbf{96.28\%} & \textbf{85.22\%} & \textbf{97.48\%} & \textbf{97.40\%} & \textbf{91.26\%} & \textbf{95.18\%} & \textbf{94.78\%} & \textbf{87.28\%} & \textbf{96.74\%} & \textbf{93.94\%} \\
   & 8/255 & 100.00\% & \textbf{86.50\%} & \textbf{78.26\%} & \textbf{49.38\%} & \textbf{81.86\%} & \textbf{82.94\%} & \textbf{61.86\%} & \textbf{73.18\%} & \textbf{74.16\%} & \textbf{51.92\%} & \textbf{81.56\%} & \textbf{72.16\%} \\
   & 4/255  & 99.92\%  & \textbf{52.32\%} & 36.94\% & 17.22\% & \textbf{44.50\%} & \textbf{46.70\%} & 23.52\% & \textbf{32.18\%} & 34.12\% & \textbf{16.66\%} & 45.76\% & \textbf{34.99\%}  \\\midrule
\multirow{3}{*}{LinBP ($\times10$) +ElasticNet}            & 16/255 & 100.00\% & 94.04\% & 88.02\% & 68.42\% & 90.82\% & 91.62\% & 80.52\% & 85.20\% & 85.68\% & 70.48\% & 90.54\% & 84.53\% \\
 & 8/255  & 99.94\%  & 71.88\% & 56.40\% & 30.74\% & 61.74\% & 65.76\% & 42.92\% & 51.82\% & 52.98\% & 30.42\% & 63.98\% & 52.86\% \\
 & 4/255  & 99.32\%  & 35.30\% & 22.02\% & 10.70\% & 27.22\% & 31.48\% & 14.00\% & 18.64\% & 20.52\% & 8.82\%  & 31.54\% & 22.02\%  \\\midrule
\multirow{3}{*}{LinBP ($\times10$) +SVR}            & 16/255 & 100.00\% & 97.54\% & 95.50\% & 84.48\% & 96.94\% & 96.86\% & 90.28\% & 94.62\% & 94.34\% & 86.22\% & 96.40\% & 93.32\% \\
& 8/255 & 100.00\% & 85.70\% & 77.40\% & 49.24\% & 81.58\% & 82.58\% & 61.40\% & 72.20\% & 73.34\% & 51.38\% & 81.20\% & 71.60\% \\
& 4/255 & 99.88\%  & 51.88\% & \textbf{37.16\%} & \textbf{17.32\%} & 44.26\% & 46.60\% & \textbf{23.62\%} & 31.78\% & \textbf{34.18\%} & 16.44\% & \textbf{46.20\%} & 34.94\%  \\\bottomrule
   \end{tabular}   
   }
 \end{center} 
\end{table*} 

\begin{table*}[t!]
 \caption{Further performance gain can be achieved by performing more than one run of a baseline method and using a more powerful baseline attack, \eg, LinBP. The attack success rates on ImageNet are reported, with $\ell_2$ constraints in the untargeted setting. The symbol * is used to indicate when the source model is used as the target. }\label{tab:linbp_l2}
 \begin{center}\resizebox{0.99\linewidth}{!}{
   \begin{tabular}{C{1.25in}C{0.35in}C{0.5in}C{0.5in}C{0.58in}C{0.5in}C{0.5in}C{0.58in}C{0.5in}C{0.5in}C{0.5in}C{0.5in}C{0.53in}C{0.45in}}
    \toprule
    Method  & $\epsilon$ & ResNet-50* & VGG-19 \cite{Simonyan2015}  & ResNet-152$\,$\cite{He2016}  & Inception v3 \cite{Szegedy2016}  & DenseNet \cite{Huang2017densely} & MobileNet v2 \cite{Sandler2018mobilenetv2} & SENet \cite{Hu2018} & ResNeXt \cite{Xie2017aggregated}  & \ WRN\ \cite{Zagoruyko2016}  & PNASNet \cite{Liu2018}  & MNASNet \cite{Tan2019mnasnet} & Average  \\
    \midrule
\multirow{3}{*}{LinBP}         & 15 &100.00\% & 96.98\% & 95.80\% & 79.96\% & 96.16\% & 94.98\% & 89.42\% & 93.96\% & 93.62\% & 82.16\% & 94.34\% & 91.74\% \\
& 10 & 100.00\% & 89.88\% & 84.00\% & 56.16\% & 85.34\% & 84.72\% & 69.46\% & 79.08\% & 79.76\% & 58.44\% & 82.82\% & 76.97\% \\
& 5 & 100.00\% & 61.36\% & 45.70\% & 21.50\% & 52.02\% & 52.32\% & 30.30\% & 39.26\% & 42.02\% & 22.26\% & 49.50\% & 41.62\% \\\midrule
\multirow{3}{*}{LinBP ($\times10$) +RR}         & 15 & 100.00\% & \textbf{98.10\%} & \textbf{96.62\%} & \textbf{88.28\%} & \textbf{97.82\%} & \textbf{97.40\%} & \textbf{92.18\%} & \textbf{95.38\%} & \textbf{95.58\%} & \textbf{89.88\%} & 97.06\% & \textbf{94.83\%} \\
                                          & 10 & 100.00\% & \textbf{93.76\%} & \textbf{89.76\%} & \textbf{68.66\%} & \textbf{92.00\%} & 92.32\% & \textbf{78.30\%} & \textbf{86.70\%} & 86.64\% & \textbf{71.44\%} & \textbf{91.10\%} & \textbf{85.07\%} \\
                                          & 5  & 99.94\%  & \textbf{72.08\%} & \textbf{58.66\%} & \textbf{31.52\%} & \textbf{64.18\%} & \textbf{66.38\%} & \textbf{41.46\%} & \textbf{52.44\%} & \textbf{54.92\%} & \textbf{33.02\%} & \textbf{65.22\%} & \textbf{53.99\%} \\\midrule
\multirow{3}{*}{LinBP ($\times10$) +ElasticNet} & 15 & 99.96\%  & 94.12\% & 89.80\% & 73.04\% & 92.12\% & 92.30\% & 80.12\% & 86.82\% & 87.12\% & 74.72\% & 91.48\% & 86.16\% \\
                                       & 10 & 99.96\%  & 84.78\% & 74.92\% & 51.22\% & 78.84\% & 81.34\% & 59.84\% & 70.28\% & 70.70\% & 50.94\% & 80.14\% & 70.30\% \\
                                       & 5  & 99.04\%  & 54.26\% & 40.42\% & 21.00\% & 46.40\% & 50.96\% & 28.52\% & 36.04\% & 37.12\% & 20.78\% & 49.76\% & 38.53\% \\\midrule
\multirow{3}{*}{LinBP ($\times10$) +SVR}        & 15 & 100.00\% & 98.04\% & 96.40\% & 87.36\% & 97.52\% & 97.34\% & 91.92\% & 95.16\% & 95.50\% & 89.12\% & \textbf{97.10\%} & 94.55\% \\
                                       & 10 & 100.00\% & 93.44\% & 89.02\% & 68.60\% & 91.34\% & \textbf{92.34\%} & 77.48\% & 85.92\% & \textbf{86.70\%} & 70.52\% & 91.02\% & 84.64\% \\
                                       & 5  & 99.90\%  & 71.28\% & 58.20\% & 31.74\% & 63.94\% & 66.70\% & 41.52\% & 52.32\% & 54.40\% & 32.86\% & 64.94\% & 53.79\%\\\bottomrule
   \end{tabular}   
   }
 \end{center} 
\end{table*} 

\begin{figure}[th]
\begin{center}
\subfloat[PGD+ElasticNet]{\label{fig:5a}
\includegraphics[width=0.31\columnwidth]{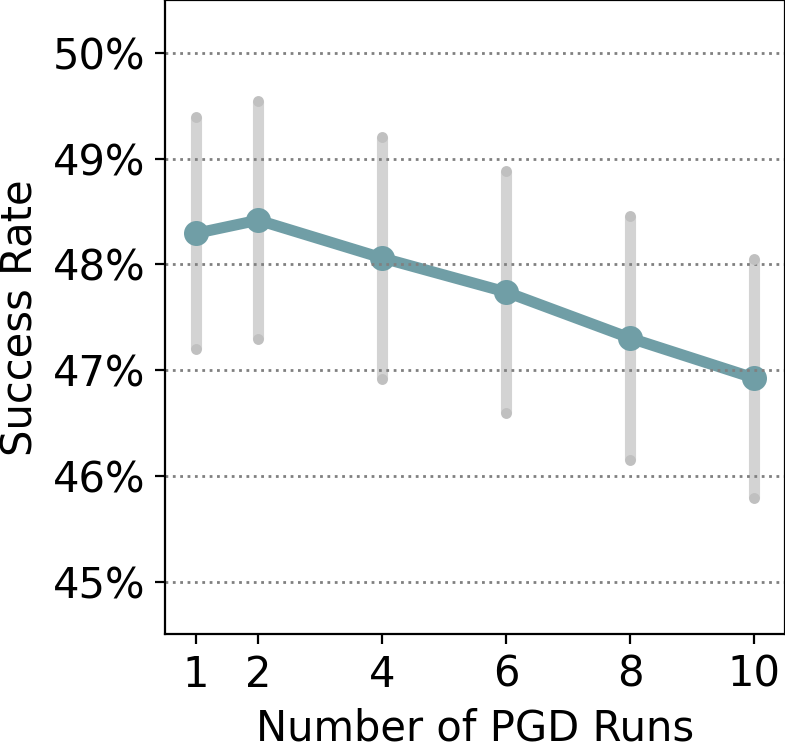}}\hskip 4pt
\subfloat[PGD+RR]{\label{fig:5b}
\includegraphics[width=0.31\columnwidth]{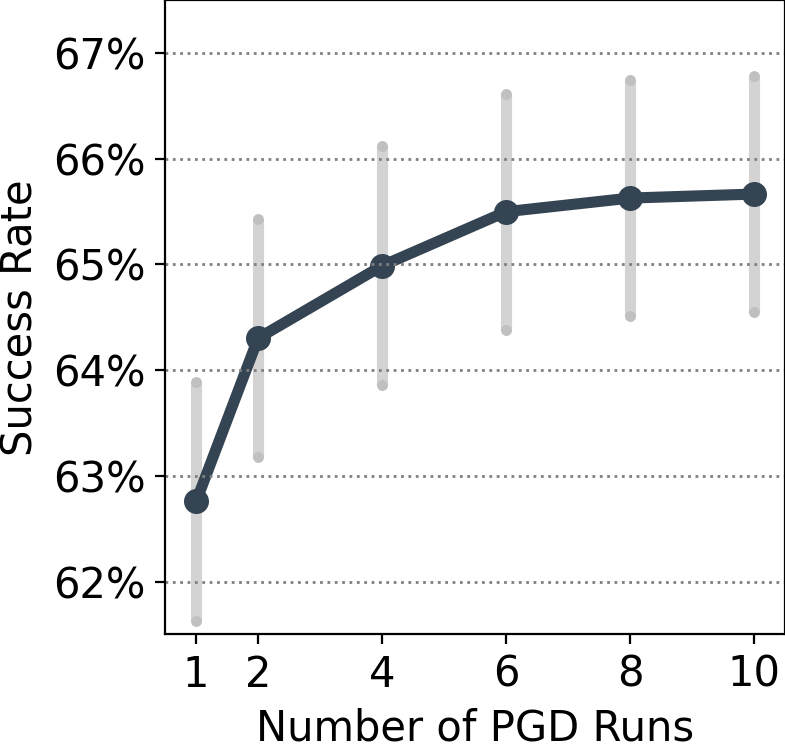}}\hskip 4pt
\subfloat[PGD+SVR]{\label{fig:5c}
\includegraphics[width=0.31\columnwidth]{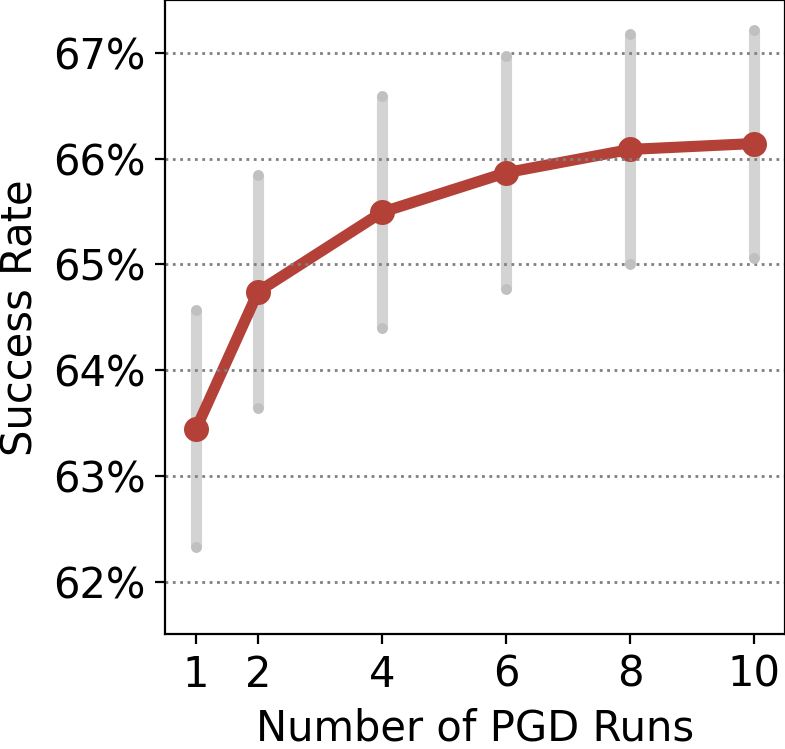}}\hskip 4pt
\caption{How the number of PGD runs affects the average attack success rate in our framework. $\epsilon=8/255$.}
\label{fig:PGD-restart}
\end{center}
\vskip -0.15in
\end{figure}

Our first attempt is to perform several runs of the PGD baseline attack~\cite{Madry2018}, which adopts a random perturbation (on the benign example) ahead of performing the standard I-FGSM. Benefiting from the random perturbations, we can expect different temporary results for different runs of the baseline attack. An experiment was conducted to evaluate the effectiveness of this simple idea. Specifically, we performed $10$ runs of PGD used RR, ElasticNet, and SVR to established the linear mapping. We obtained an average success rate of $66.14\%$ on ImageNet with SVR under $\epsilon=8/255$, outperforming the result in Table~\ref{tab:linear_regression} significantly. How the performance of our methods varies with the number of PGD runs is illustrated in Figure~\ref{fig:PGD-restart}. It can be seen that, for RR and SVR, increasing the number of PGD runs improves the average attack success rate. While for ElasticNet which is a sparse coding method, the result is unstable. 
For $\ell_\infty$ attacks with other $\epsilon$ settings (\ie, $16/255$ and $4/255$) and $\ell_2$ attacks, similar observations can be made, \ie, the best results is obtained with SVR, and RR achieved the second best.

\begin{figure}[th]
\begin{center}
\subfloat[LinBP+ElasticNet]{\label{fig:6a}
\includegraphics[width=0.31\columnwidth]{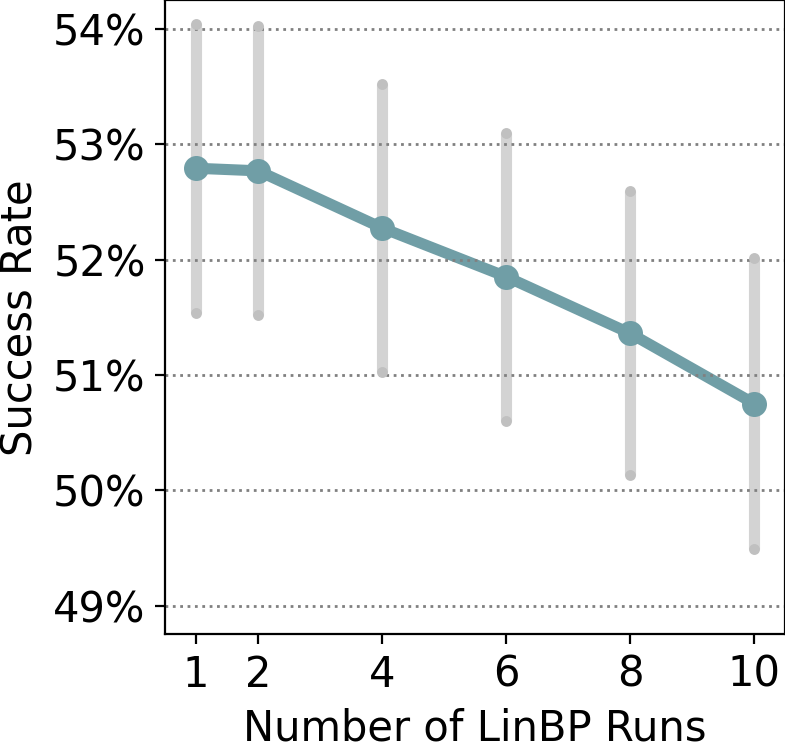}}\hskip 4pt
\subfloat[LinBP+RR]{\label{fig:6b}
\includegraphics[width=0.31\columnwidth]{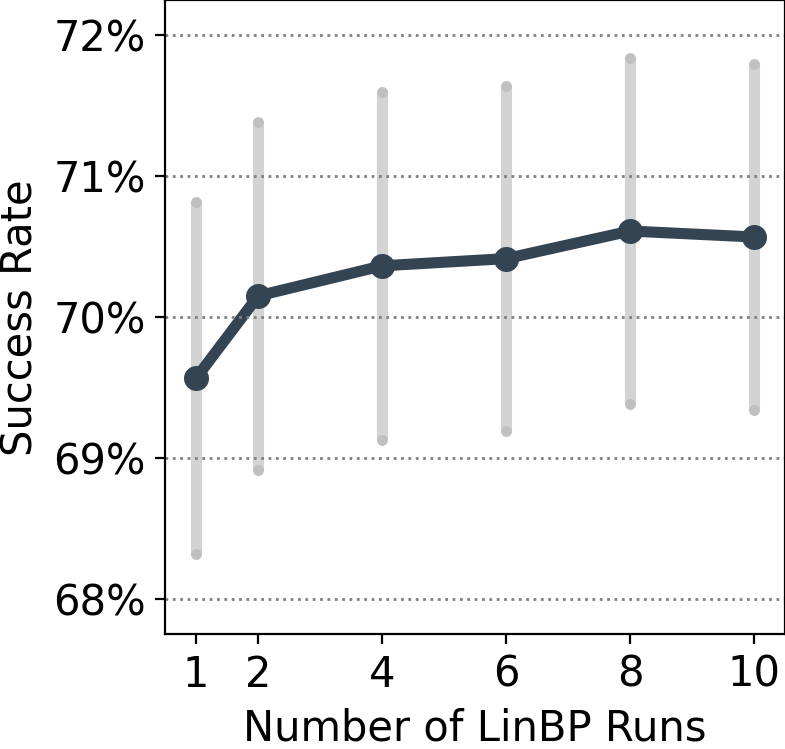}}\hskip 4pt
\subfloat[LinBP+SVR]{\label{fig:6c}
\includegraphics[width=0.31\columnwidth]{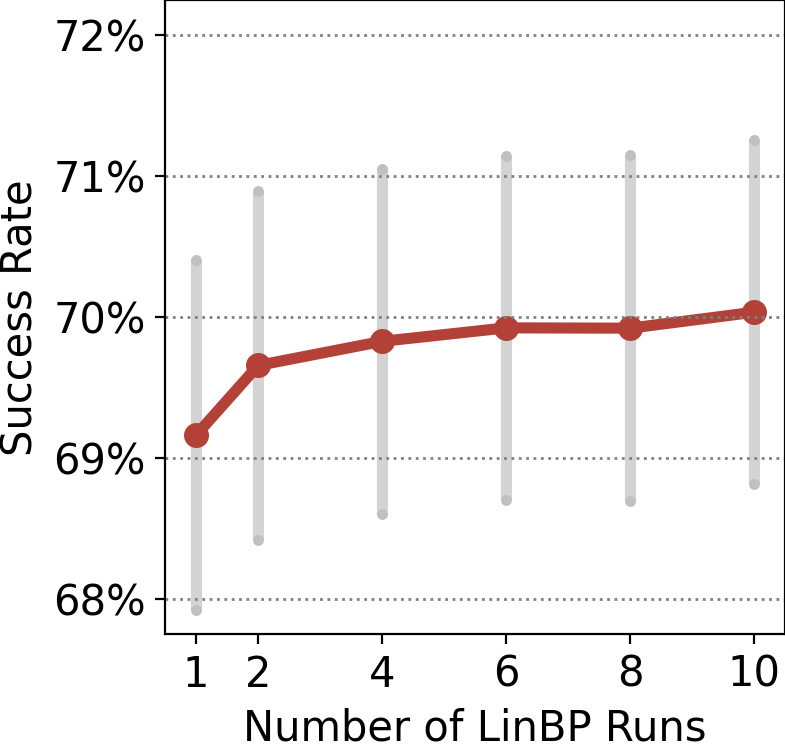}}\hskip 4pt
\caption{How the number of LinBP runs affects the average attack success rate in our framework. $\epsilon=8/255$.}
\label{fig:LinBP-restart}
\end{center}
\vskip -0.25in
\end{figure}

More powerful baseline attack also leads to more transferable adversarial examples after refinement in our framework. Therefore, it is natural to try more advanced methods than I-FGSM and PGD. We have tested with MI-FGSM~\cite{Dong2018}, TAP~\cite{Zhou2018}, and LinBP~\cite{guo2020back}. Together with random perturbations beforehand, each of these methods can be performed more than once for collecting diverse directional guides. Results showed that LinBP is the best in this regard, assisting RR in our framework to achieve an average attack success rate of $94.83\%$, $72.16\%$, and $53.99\%$ under $\epsilon=16/255$, $8/255$, and $4/255$, respectively. For comparison, our ILA++ in its default setting achieves $88.57\%$, $63.38\%$, and $30.59\%$, respectively (as reported in Table~\ref{tab:imagenet}). Figure~\ref{fig:LinBP-restart}, Table~\ref{tab:linbp_linf}, and Table~\ref{tab:linbp_l2} report more detailed results. Due to the space limit of the paper, we omit results with MI-FGSM and TAP in this paper, and it is worthy noting that different baseline methods might favor different linear regression models, cf. Figure~\ref{fig:PGD-restart} (in which +SVR is the best) and Figure~\ref{fig:LinBP-restart} (in which +RR is the best).

\section{Detailed Experimental Settings}\label{sec:exp}

In this section, we will introduce detailed experimental settings for all the experiments in this paper.

\subsection{General Settings}
All the experiments were performed on the same set of ImageNet~\cite{Russakovsky2015} models, \ie, ResNet-50~\cite{He2016}, VGG-19~\cite{Simonyan2015}, ResNet-152~\cite{He2016}, Inception v3~\cite{Szegedy2016}, DenseNet~\cite{Huang2017densely}, MobileNet v2~\cite{Sandler2018mobilenetv2}, SENet~\cite{Hu2018}, ResNeXt~\cite{Xie2017aggregated}, WRN~\cite{Zagoruyko2016}, PNASNet~\cite{Liu2018}, and MNASNet~\cite{Tan2019mnasnet}. These models are popularly used and can achieve reasonably high prediction accuracy on the ImageNet validation set: $76.15\%$ (ResNet-50), $74.24\%$ (VGG-19), $78.31\%$ (ResNet-152), $77.45\%$ (Inception v3), $74.65\%$ (DenseNet), $71.18\%$ (MobileNet v2), $81.30\%$ (SENet), $79.31\%$ (ResNeXt), $78.84\%$ (WRN), $82.74\%$ (PNASNet), and $73.51\%$ (MNASNet). To keep in line with the work in~\cite{li2020yet}, we chose ResNet-50 as the source model. Some victim models have different pre-processing pipelines, and we followed their official settings. For instance, for DenseNet, we resized its input images to $256\times256$ and then cropped them to $224\times224$ at center. 

We randomly sampled 5,000 test images that can be correctly classified by all the victim models from the ImageNet validation set for evaluating attacks. Specifically, these images were from 500 (randomly chosen) classes, and we had 10 images per class.

For the constraint on perturbations and step size, we tested $\epsilon = 16/255, 8/255, 4/255$ with a common step size of $1/255$ for $\ell_\infty$ attacks, and we tested $\epsilon = 15.0, 10.0, 5.0$ using $\epsilon/5$ as the step size for $\ell_2$ attacks. We run $100$ iterations for I-FGSM and LinBP. We let $p=10$ for attacks using temporary results from the baseline attack (in Section~\ref{sec:linear}). Then, with $10$ runs in Section~\ref{sec:diverse}, we have $10\times 10=100$ samples for establishing a linear mapping. After updating the adversarial examples at each attack iteration, they were clipped to the range of [0.0, 1.0] to keep the validity of being images. 

Since the concerned attacks utilize intermediate-level representations, the choice of hidden layer should have an impact on the performance of the attacks.
Unless otherwise clarified, we always chose the first block of the third meta-block of ResNet-50. For LinBP, the last six building blocks in ResNet-50 were modified to be more linear during backpropagation.

\section{Conclusion}
We have demonstrated that the transferability of adversarial examples can be substantially improved by establishing a linear mapping directly from some middle-layer features (of a source DNN) to its prediction loss. Various linear regression models can thus be chosen for achieving the goal, and we have shown the effectiveness using RR, ElasticNet, and SVR. We have carefully analyzed core components of such a framework and shown that, given powerful directional guides, the magnitude of intermediate-level feature discrepancies is correlated with the transferability. Using random directional guides or normalizing feature discrepancies as in Eq.~\eqref{eq:opt2} shall lead to poor attack performance. Given all these facts, we have demonstrated that further improved attack performance can be obtained via collecting more diverse yet still powerful directional guides. New state-of-the-arts have been achieved.


\bibliographystyle{IEEEtran}
\bibliography{ref}

\begin{IEEEbiography}[{\includegraphics[width=1in,height=1.25in,clip,keepaspectratio]{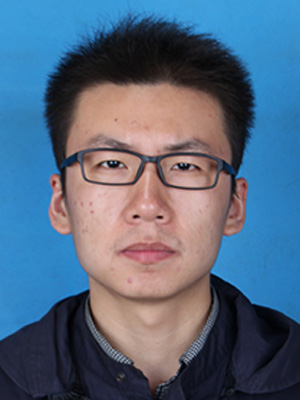}}]{Yiwen Guo}
received the B.E. degree from Wuhan University in 2011, and the Ph.D. degree from Tsinghua University in 2016. He was a research scientist at ByteDance AI Lab, Beijing. Prior to this, he was a staff research scientist at Intel Labs China. His current research interests include computer vision, pattern recognition, and machine learning.
\end{IEEEbiography}

\begin{IEEEbiography}[{\includegraphics[width=1in,height=1.25in,clip,keepaspectratio]{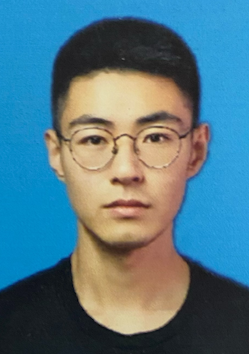}}]{Qizhang Li}
received the B.E. and M.E. degrees in 2018 and 2021, respectively, from Harbin Engineering University, Harbin, China. He is currently pursuing the Ph.D. degree in computer science and technology with the School of Computer Science and Technology, Harbin Institute of Technology, Harbin, China. His current research interests include computer vision and machine learning.
\end{IEEEbiography}

\begin{IEEEbiography}[{\includegraphics[width=1in,height=1.25in,clip,keepaspectratio]{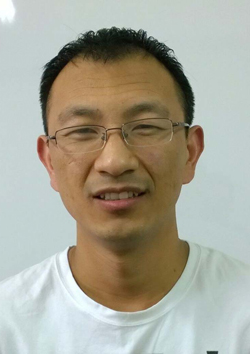}}]{Wangmeng Zuo} (M'09, SM'14)  received the Ph.D. degree in computer application technology from the Harbin Institute of Technology, Harbin, China, in 2007.
He is currently a Professor in the School of Computer Science and Technology, Harbin Institute of Technology. His current research interests include image enhancement and restoration, image and face editing, object detection, visual tracking, and image classification. He has published over 100 papers in top tier academic journals and conferences. He has served as a Tutorial Organizer in ECCV 2016, an Associate Editor of \emph{IEEE Trans. Pattern Analysis and Machine Intelligence}, \emph{The Visual Computers}, \emph{Journal of Electronic Imaging}.
\end{IEEEbiography}

\begin{IEEEbiography}[{\includegraphics[width=1in,height=1.25in,clip,keepaspectratio]{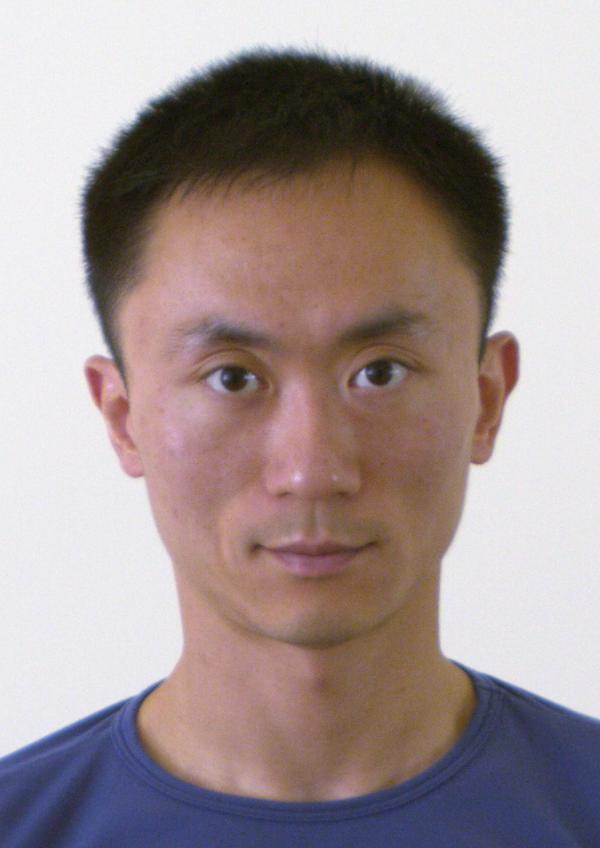}}]{Hao Chen}
is a professor at the Department of Computer Science at the University of California, Davis. He received his PhD at the Computer Science Division at the University of California, Berkeley, and his BS and MS from Southeast University. His research interests are computer security, machine learning, and mobile computing.
\end{IEEEbiography}

\newpage

\begin{appendices}
\section{Approximation to ILA++} \label{sec:app}
This section is a brief recap of the derivation in our ECCV paper~\cite{li2020yet}, based on which an approximation to the RR result is given.

We have $\mathbf w^\ast = (\mathbf H^T \mathbf H + \lambda \mathbf I_m)^{-1} \mathbf H^T \mathbf r$ for RR, where the inverse of an $m\times m$ matrix $\mathbf H^T \mathbf H + \lambda \mathbf I_m$ is required and $m$, the dimensionality of the middle-level DNN feature, is high. Therefore, it is computationally very demanding to calculate $\mathbf w^\ast$ using such a formulation directly. To reduce the computational burden, we proposed to use the Woodbury identity
\begin{equation}\label{eq:woodbury}
\begin{aligned}
    \mathbf H^T \mathbf H + \lambda \mathbf I_m &= \frac{1}{\lambda}I - \frac{1}{\lambda^2}\mathbf H^T(\frac{1}{\lambda}\mathbf H \mathbf H^T + \mathbf I_p)^{-1} \mathbf H \\
    &=\frac{1}{\lambda}I - \frac{1}{\lambda}\mathbf H^T(\mathbf H \mathbf H^T + \lambda\mathbf I_p)^{-1} \mathbf H
\end{aligned}
\end{equation}
such that the inverse can be calculated on a much smaller matrix. We further notice $\mathbf H^T(\mathbf H \mathbf H^T + \lambda\mathbf I)^{-1} \mathbf H\approx \mathbf 0$ with a strong regularization, which leads to Eq.~\eqref{eq:opt0_app}.

Although such an approximation can be obtained for RR, in order to perform fair comparisons, most experimental results reported in this paper were still obtained using Eq.~\eqref{eq:opt0} directly for RR. 

\end{appendices}

\end{document}